\documentclass{article}

\def \TRtitle{A Unifying Framework for Typical Multi-Task Multiple Kernel Learning Problems}
\def \TRauthors{Cong Li, Michael Georgiopoulos and Georgios C. Anagnostopoulos}
\def \TRemails{congli@eecs.ucf.edu, michaelg@ucf.edu and georgio@fit.edu}
\def \TRkeywords{Multiple Kernel Learning, Multi-task Learning, Support Vector Machines}

\usepackage{./options}
\usepackage{./packages}
\usepackage{./macros}
\begin{document}

% Make title pages
\maketitle

% Do not change. %
\ifMakeReviewDraft
	\linenumbers
\fi

\begin{abstract}
Over the past few years, \ac{MKL} has received significant attention among data-driven feature selection techniques in the context of kernel-based learning. \ac{MKL} formulations have been devised and solved for a broad spectrum of machine learning problems, including \ac{MTL}. Solving different \ac{MKL} formulations usually involves designing algorithms that are tailored to the problem at hand, which is, typically, a non-trivial accomplishment.

In this paper we present a general \ac{MT-MKL} framework that subsumes well-known \ac{MT-MKL} formulations, as well as several important \ac{MKL} approaches on single-task problems. We then derive a simple algorithm that can solve the unifying framework. To demonstrate the flexibility of the proposed framework, we formulate a new learning problem, namely \ac{PSCS} \ac{MT-MKL}, and demonstrate its merits through experimentation.
\end{abstract}

% Do not change. %

% Input sections from separate files. Modify these fields as necessary.
% reset all acronyms
\acresetall

%%%%%%%%%%%%%%%%%%%%%%%%%%%%%%%%%%%%%%%%%%%%%%%%%%%%%%%%%%%%%%%%%%%%%%%%%%%%%%%%
%%%%%%%%%%%%%%%%%%%%%%%%%%%%%%%%%%%%%%%%%%%%%%%%%%%%%%%%%%%%%%%%%%%%%%%%%%%%%%%%
%%%%%%%%%%%%%%%%%%%%%%%%%%%%%%%%%%%%%%%%%%%%%%%%%%%%%%%%%%%%%%%%%%%%%%%%%%%%%%%%
\section{Introduction}
\label{sec:Introduction}

Kernel methods play an important role in machine learning due to their elegant property of implicitly mapping samples from the original space into a potentially infinite-dimensional feature space, in which inner products can be calculated directly via a kernel function. It is desired that samples are appropriately distributed in the feature space, such that kernel-based models, like \acp{SVM}\cite{Cortes1995} \cite{Vapnik1998}, could perform better in the feature space than in the original space. Since the feature space is implicitly defined via the kernel function, it is important to choose the kernel appropriately for a given task. In recent years, substantial effort has been devoted on how to learn such kernel function (or equivalently, kernel matrix) from the available data. According to the \ac{MKL} approach \cite{Lanckriet2004}, which is one of the most popular strategies for learning kernels, multiple predetermined kernels are, most commonly, linearly combined. Subsequently, in lieu of tuning kernel parameters via some validation scheme, the combination coefficients are adapted to yield the optimal kernel in a data-driven manner. A thorough review of MKL methods and associated algorithms is provided in \cite{Gonen2011}. 

So far, several \ac{MKL} formulations, along with their optimization algorithms, have been proposed. The earlier work in \cite{Lanckriet2004} suggests a \ac{MKL} formulation with trace constraints over the linearly combined kernels, which is further transformed into a solvable Semi-Definite Programming problem. Also, Simple-MKL \cite{Rakotomamonjy2008} relies on a \ac{MKL} formulation with $L_1$-norm constrained coefficients. In each iteration of the algorithm, a \ac{SVM} problem is solved by taking advantage of an existing efficient \ac{SVM} solver and the coefficients are updated via a gradient-based scheme. A similar algorithm is also applied in \cite{Xu2013}. In \cite{Kloft2008}, an $L_2$-norm constraint is applied to the linear combination coefficients and the proposed min-max formulation is transformed into a \ac{SIP} problem, which is then solved via a cutting plane algorithm. Moreover, two algorithms are proposed in \cite{Kloft2009} to solve the $L_p$-\ac{MKL} formulation, where the coefficient constraint is generalized to an $L_p$-norm constraint. The relationship between the latter $L_p$-\ac{MKL} formulation and one that entails a Group-Lasso regularizer is pointed out in \cite{Xu2010}, which utilizes a block coordinate descent algorithm to solve the problem in the primal domain. Besides \ac{MKL}-based \ac{SVM} models, a \ac{MKL} formulation for \ac{KRR}\cite{Saunders1998} has been proposed in \cite{Cortes2009a}. Cast as a min-max problem, it is solved via an interpolated iterative algorithm, which takes advantage of the closed-form solution for the \ac{KRR} kernel coefficients. %of \ac{KRR} and the kernel coefficients. 

Furthermore, \ac{MKL} has been applied to \ac{MTL} giving rise to \ac{MT-MKL} approaches, where several tasks with shared feature representations are simultaneously learned. As was the case with single-task \ac{MKL} approaches mentioned in the previous paragraph, most \ac{MT-MKL} formulations that have appeared in the literature are accompanied by an algorithm tailored to their particular formulation. For example, a framework is formulated in \cite{Tang2009}, where individual tasks utilize their own task-specific space in conjunction with a shared space component, while the balance between these two components is controlled through weights. Subsequently, an exchange algorithm is proposed to solve the resulting min-max problem. Additionally, in \cite{Aflalo2011}, a Group-Lasso regularizer on \ac{SVM} weights is considered that yields coefficient sparsity within a group of tasks and non-sparse regularization across groups. The resulting problem is solved in the dual domain via a mirror descent-based algorithm. Also, in \cite{Rakotomamonjy2011}, the Group-Lasso regularizer is generalized to $L_p-L_q$ regularization and two optimization problems are addressed, one being convex and the other non-convex. Next, for each of these two formulations, a specialized algorithm is proposed. Furthermore, in \cite{Widmer2010}, \ac{MT-MKL} is formulated in the primal domain by penalizing the \ac{SVM} weight of each task to be close to a shared common weight. Finally, maximum entropy discrimination is employed in \cite{Jebara2011} to construct a \ac{MT-MKL} framework.

Proposing useful \ac{MKL} formulations and devising effective algorithms specifically tailored to solving them is, in most cases, a non-trivial task; such an endeavor requires fair amounts of insight and ingenuity. This is even more so true, when one considers applying \ac{MKL} to \ac{MTL} problems. In this paper, we present a general kernel-based \ac{MT-MKL} framework that subsumes well-known \ac{MT-MKL} formulations, as well as several important \ac{MKL} approaches on single-task machine learning problems. Its unifying character stems from its applicability to prominent kernel-based tasks, such as \ac{SVM}-based binary classification, \ac{KRR}-based regression and outlier detection based on One-Class \ac{SVM} \cite{Scholkopf2001} and \ac{SVDD} \cite{Tax1999}, to name a few important ones. Also, it accommodates various feature space sharing approaches that may be encountered in a \ac{MT-MKL} setting, as well as single-task \ac{MKL} settings with different constraints. For example, it subsumes $L_2$-MKL and $L_p$-MKL considered in \cite{Kloft2008} and \cite{Kloft2009}, the KRR-based \ac{MKL} in \cite{Cortes2009a} and the \ac{MT-MKL} formulations in \cite{Tang2009} and \cite{Rakotomamonjy2011}. We introduce our framework in \sref{sec:ProblemFormulation}.

In \sref{sec:ExactPenaltyFunctionMethod} we state an equivalency between solving \ac{SIP} problems and \ac{EPF} optimization, which may be viewed as a useful result on its own. As the new framework can be cast as a \ac{SIP} problem, the previous equivalency allows us in \sref{sec:Algorithm} to eventually derive a straightforward and easy to implement algorithm for solving the new EPF-based formulation. For a given \ac{MT-MKL} problem that is a special case of our framework, the availability of our algorithm eliminates the need of using an existing, potentially complicated, algorithm or deriving a new algorithm that is tailored to the problem at hand. A further major advantage of the algorithm is that it can leverage from already existing efficient kernel machine solvers (\eg, SVM solvers) or closed-form solutions (\eg, KRR solution) for given problems. 

In \sref{sec:PSCS_model}, we present a new specialization of our framework, called \ac{PSCS} \ac{MT-MKL}. It permits some tasks to share a common feature space, while other tasks are allowed to utilize their own feature space. This approach follows the spirit of recent \ac{MTL} research, which allows tasks to be grouped, while tasks belonging to the same group can share information with each other. Some examples include the works in \cite{Xue2007}, \cite{Jacob2008}, \cite{Kang2011} and \cite{Zhong2012}, to name a few. This is a generalization of the traditional \ac{MTL} setting, according to which all tasks are considered as one group and are learned concurrently. Our \ac{PSCS} \ac{MT-MKL} formulation is a combination of the grouped \ac{MTL} setting and \ac{MKL}, which appreciably differs from previously cited works. The new formulation concretely showcases the generality of the proposed \ac{MT-MKL} framework, the flexibility in choosing a suitable \ac{MT-MKL} feature sharing strategy, as well as the usefulness of the derived algorithm. The merits of \ac{PSCS} are illustrated in \sref{sec:ExperimentalResults}, where we compare its performance on classification benchmark data sets against the common-space (CS) and independent-space (IS) alternatives. Proofs of our analytical work are given in the appendices.

\section{Problem Formulation}
\label{sec:ProblemFormulation}

Consider a supervised learning task with parameter $\boldsymbol{\alpha}$, which can be expressed in the form

\begin{equation}
	\max_{\boldsymbol{\alpha}\in\boldsymbol{\Omega} ( \boldsymbol{\alpha} )} \bar{g}(\boldsymbol{\alpha}, \boldsymbol{K})
	\label{eq:one}
\end{equation}

\noindent where $\boldsymbol{K}$ is the kernel matrix of the training data set, \ie, its $(i,j)$ entry is $K ( \boldsymbol{x}_i,\boldsymbol{x}_j ) \triangleq \langle \phi( \boldsymbol{x}_i ),\phi ( \boldsymbol{x}_j )\rangle_{\mathcal{H}}$, and $\mathcal{H}$ is the Hilbert space reproduced by the kernel function $k$. Here, $\phi$ is the feature mapping that is implied by the kernel function $k (\cdot, \cdot )$ and $\boldsymbol{x}_i, \boldsymbol{x}_j \in \mathcal{X}$, where $\mathcal{X}$ is an input set. Also, suppose $\bar{g}$ has a finite maximum, a finite number of local maxima with respect to $\boldsymbol{\alpha}$ in the feasible set $\boldsymbol{\Omega}( \boldsymbol{\alpha}) \subset \mathbb{R}^n$ and is affine with respect to the individual entries of $\boldsymbol{K}$. Some well-known supervised learning tasks that feature these characteristics are considered in \ac{SVM}, \ac{KRR}, \ac{SVDD} %\cite{Tax1999} 
and One-Class \ac{SVM}. %\cite{Scholkopf2001}
Their dual-domain objective functions $\bar{g}$ for $n$ training samples are given in \eref{eq:gBarSVM} through \eref{eq:gBar1CSVM} respectively.

\begin{flalign}
	\label{eq:gBarSVM} 
	&\bar{g}_{\scriptscriptstyle SVM} ( \boldsymbol{\alpha}, \boldsymbol{K} ) \triangleq \boldsymbol{\alpha}' \boldsymbol{1} - \frac{1}{2} \boldsymbol{\alpha}' \boldsymbol{YKY} \boldsymbol{\alpha}, &\\
	& \text{with} \; \; \boldsymbol{\Omega} ( \boldsymbol{\alpha} ) \triangleq \{ \boldsymbol{\alpha} \in \mathbb{R}^n: \boldsymbol{0} \preceq \boldsymbol{\alpha} \preceq C \boldsymbol{1}, \boldsymbol{\alpha}' \boldsymbol{y} = 0 \} & \notag
\end{flalign}
\begin{flalign}
	\label{eq:gBarKRR}
	&\bar{g}_{\scriptscriptstyle KRR} ( \boldsymbol{\alpha}, \boldsymbol{K} ) \triangleq 2 \boldsymbol{\alpha}' \boldsymbol{y} - \boldsymbol{\alpha}' ( \lambda \boldsymbol{I} + \boldsymbol{K} ) \boldsymbol{\alpha},& \\
	& \text{with} \; \; \boldsymbol{\Omega} ( \boldsymbol{\alpha} ) \triangleq \mathbb{R}^n & \notag% There is no constraint for KRR.
\end{flalign}
\begin{flalign}
	\label{eq:gBarSVDD}
	&\bar{g}_{\scriptscriptstyle SVDD} ( \boldsymbol{\alpha}, \boldsymbol{K} ) \triangleq \boldsymbol{\alpha}' \boldsymbol{k} - \boldsymbol{\alpha}' \boldsymbol{K} \boldsymbol{\alpha}, & \\
	& \text{with} \; \;\boldsymbol{\Omega} ( \boldsymbol{\alpha} ) \triangleq \{ \boldsymbol{\alpha} \in \mathbb{R}^n: \boldsymbol{0} \preceq \boldsymbol{\alpha} \preceq C \boldsymbol{1}, \boldsymbol{\alpha}' \boldsymbol{1} = 1 \} & \notag
\end{flalign}
\begin{flalign}
\label{eq:gBar1CSVM}
	&\bar{g}_{\scriptscriptstyle OCSVM} ( \boldsymbol{\alpha}, \boldsymbol{K} ) \triangleq - \boldsymbol{\alpha}' \boldsymbol{K\alpha},& \\
	&\text{with} \; \; \boldsymbol{\Omega} ( \boldsymbol{\alpha} ) \triangleq \{ \boldsymbol{\alpha} \in \mathbb{R}^n: \boldsymbol{0} \preceq \boldsymbol{\alpha} \preceq \frac{1}{\nu l} \boldsymbol{1}, \boldsymbol{\alpha}' \boldsymbol{1} = 1 \} & \notag
\end{flalign}

\noindent
In the above examples, $\cdot'$ signifies transposition vector/matrix , $\boldsymbol{y} \triangleq [ y_1, \cdots, y_n ]'$ is the target vector, $\boldsymbol{k} \triangleq [ k ( \boldsymbol{x}_1, \boldsymbol{x}_1 ),\cdots,k ( \boldsymbol{x}_n, \boldsymbol{x}_n ) ]'$ and $\boldsymbol{Y} \triangleq \mathrm{diag}(\boldsymbol{y})$, where $\mathrm{diag}(\cdot)$ is an operator yielding a diagonal matrix, whose diagonal is formed by the operand's vector argument. Additionally, $C, \lambda, \nu, l$ are scalar non-negative regularization parameters for the corresponding learning problems and $\boldsymbol{1}$ is the all-ones vector of appropriate dimension. 

Assume, now, that we have $T$ such tasks and let $ \{ \boldsymbol{x}_i^t, y_i^t \}$, $i=1,\cdots , N_t$, be the training data for task $t$, where $t=1, \cdots, T$. Let $k_m ( \boldsymbol{x}_i^t,\boldsymbol{x}_j^t ) \triangleq \langle \phi_m ( \boldsymbol{x}_i^t ),\phi_m ( \boldsymbol{x}_j^t ) \rangle_{\mathcal{H}_m}$, where $m=1,\cdots M$, be the $m$-th pre-specified kernel function and $\phi_m$ be the $m$-th associated feature mapping. Also, let $\boldsymbol{K}_m^t$ be the $N_t \times N_t$ kernel matrix of the training data corresponding to the $t$-th task and calculated using the $m$-th kernel function, \ie\ $\boldsymbol{K}_m^t$ has elements $k_m ( \boldsymbol{x}_i^t,\boldsymbol{x}_j^t )$, where $i, j = 1, \cdots, N_t$. In this paper we consider the \ac{MT-MKL} framework that is formulated as follows:

\begin{equation}
	\min_{\boldsymbol{\theta}\in \boldsymbol{\Psi} ( \boldsymbol{\theta} )} \max_{\boldsymbol{a} \in \boldsymbol{\Omega} ( \boldsymbol{a} )} \sum_{t=1}^T \bar{g}(\boldsymbol{\alpha}^t, \sum_{m=1}^M \theta_m^t \boldsymbol{K}_m^t)
	\label{eq:framework}
\end{equation}

\noindent where $\boldsymbol{a} \triangleq [\boldsymbol{\alpha}^{1'},\cdots,\boldsymbol{\alpha}^{T'}]^{'} \in \mathbb{R}^N$, $N = \sum_{t=1}^T N_t$, and $\boldsymbol{\alpha}^{t'}$ denotes the transpose of $\boldsymbol{\alpha}^{t} \in \mathbb{R}^{N_t}, t=1,\cdots,T$. Similarly, $\boldsymbol{\theta}$ stands for the vector with elements $\theta_m^t$.

The framework is able to incorporate various feature space sharing approaches by appropriately specifying the feasible region of the kernel combination coefficients $\boldsymbol{\theta}$. For example, for $T=1$ it can specialize to $L_p$-norm \ac{MKL} \cite{Kloft2009}, where $p \geq 1$, by using $\boldsymbol{\Psi} ( \boldsymbol{\theta} ) = \{ \boldsymbol{\theta}:  \| \boldsymbol{\theta} \|_p \leq 1, \boldsymbol{\theta}\succeq \boldsymbol{0} \}$. Obviously, the $L_2$-norm \ac{MKL} \cite{Kloft2008} is also covered by our framework. Additionally, it can express the \ac{MT-MKL} model in \cite{Tang2009} that allows individual tasks to utilize their own task-specific space in conjunction with a shared space component. This is achieved by specifying $\boldsymbol{\Psi} ( \boldsymbol{\theta} ) =  \{ \boldsymbol{\theta}: \theta_m^t=\zeta_m + \gamma_m^t, \zeta_m \geq 0, \gamma_m^t \geq 0, \sum_{m=1}^M \theta_m^t = 1,$ $\sum_{m=1}^M \sum_{t=1}^T \gamma_m^t \leq \beta\}$. Hence, our framework also subsumes the Common Space (CS) and Independent Space (IS) \ac{MT-MKL} models as special cases. The former is obtained by letting $\theta_m^t = \zeta_m, \forall t, m$. In this case, all tasks share a common kernel function determined by the coefficients $\zeta_m$'s. Appropriate constraints can be added on $\boldsymbol{\zeta}$, such as the $L_p$-norm constraint. On the other hand, if we let $\boldsymbol{\theta}^t \triangleq [\theta_1^t,\cdots,\theta_M^t ]'$ and add independent constraints on each $\boldsymbol{\theta}^t$, then we obtain the IS model, where each task occupies its own feature space. 

To mention a final example, a Group-Lasso type regularizer is employed on the \ac{SVM} primal-domain weights in \cite{Aflalo2011}, which leads to intricate optimization problems and algorithms. Our framework specializes to this problem by specifying $\boldsymbol{\Psi}( \boldsymbol{\theta}) = \{ \boldsymbol{\theta}: \boldsymbol{\theta} \succeq  \boldsymbol{0}, (\sum_{t=1}^T \| \boldsymbol{\theta}^t \|_{p}^q )^{1/q} \leq a , p \geq 1, q \geq 1 \}$. By appropriately choosing the values of $p$ and $q$, different level of group-wise and intra-group sparsity can be obtained. In this case too, using our framework leads to an easier formulation that can be solved in a much more straightforward fashion via our algorithm.

In the next section, we first transform \pref{eq:framework} to an equivalent \ac{SIP} problem. Subsequently, we demonstrate the equivalency between general \ac{SIP} problems and \ac{EPF}-based problems. The latter result will allow us to cast \pref{eq:framework} as an \ac{EPF}-based problem, which, as it turns out, can be easily solved.

\section{Exact Penalty Function Method}
\label{sec:ExactPenaltyFunctionMethod}

In general, the min-max \pref{eq:framework} is not easy to solve. However, it can be transformed to an equivalent epigraph problem in the following \ac{SIP} form:

\begin{equation}
	\begin{aligned}
		& \min_{\omega \in \mathbb{R}, \boldsymbol{\theta}\in \boldsymbol{\Psi} ( \boldsymbol{\theta} )} \omega \\
& \text{s.t.} \; \sum_{t=1}^T \bar{g}(\boldsymbol{\alpha}^t, \sum_{m=1}^M \theta_m^t \boldsymbol{K}_m^t) \leq \omega, \; \forall \boldsymbol{a} \in\boldsymbol{\Omega}  ( \boldsymbol{a} ).
	\end{aligned}
	\label{eq:equivalent_SIP}
\end{equation}

Before solving \pref{eq:equivalent_SIP}, we first show an equivalence between \ac{EPF}-based problems \cite{Watson1981} and general \ac{SIP} problems. This will eventually facilitate the development of an algorithm for solving \pref{eq:equivalent_SIP}.

\noindent
\textbf{General \ac{SIP} problem} 

Consider the general \ac{SIP} problem

\begin{equation}
\begin{aligned}
&  \min_{\boldsymbol{x}\in \mathbb{R}^n } f(\boldsymbol{x}) \\
& \text{s.t.} \; g ( \boldsymbol{a}, \boldsymbol{x} ) \leq 0, \; \forall \boldsymbol{a} \in \boldsymbol{\Omega} ( \boldsymbol{a} );\\
& \;\;\;\;\;l_u ( \boldsymbol{x} ) = 0, u = 1,\cdots,U;  \\
& \;\;\;\;\;r_v( \boldsymbol{x} ) \leq 0, v = 1,\cdots,V.
\end{aligned}
\label{eq:general_SIP}
\end{equation}

\noindent 
with $U$ equality and $V$ inequality constraints. Now, suppose $f$, $g$, the $l_u$'s and $r_v$'s are continuously differentiable. The feasible region of $\boldsymbol{x}$ is determined by both the \ac{SIP} constraint involving $g$ and the regular equality ($l_u$'s) and inequality ($r_v$'s) constraints. It is not difficult to see that \pref{eq:equivalent_SIP} is a special case of \pref{eq:general_SIP} by defining $\boldsymbol{x} \triangleq [ \omega, \boldsymbol{\theta}' ]'$, $\boldsymbol{a} \triangleq [ \boldsymbol{\alpha}^{1'}, \cdots, \boldsymbol{\alpha}^{T'}  ]'$, $f ( \boldsymbol{x} ) \triangleq \omega$, $g ( \boldsymbol{a}, \boldsymbol{x} ) \triangleq \sum_{t=1}^T \bar{g}(\boldsymbol{\alpha}^t, \sum_{m=1}^M \theta_m^t \boldsymbol{K}_m^t) - \omega$, and letting the constraints $l_u$'s and $r_v$'s define $\boldsymbol{\Psi} ( \boldsymbol{\theta} )$.

\noindent
\textbf{\ac{EPF}-based problem} 

For fixed $\boldsymbol{x}$, assume that there are $N(\boldsymbol{x})$ local maxima $\boldsymbol{a}_i^* \in \boldsymbol{\Omega} (\boldsymbol{a})$ of $g( \boldsymbol{a}, \boldsymbol{x} )$ that satisfy $g( \boldsymbol{a}^*_i, \boldsymbol{x} ) \geq -\eta$ for some $\eta > 0$ and $g ( \boldsymbol{a}^*_i, \boldsymbol{x} ) \neq g( \boldsymbol{a}^*_j, \boldsymbol{x}), \forall \boldsymbol{a}^*_i, \boldsymbol{a}^*_j \in \Omega (\boldsymbol{a})$. Denote this set of local maxima as $E ( \boldsymbol{x} ) \triangleq \{ \boldsymbol{a}_i^* \}_{i=1}^{N( \boldsymbol{x})}$, and let $I(\boldsymbol{x}) \triangleq \{1, 2, \ldots, N(\boldsymbol{x}) \}$. Obviously, each $\boldsymbol{a}_i^*$ depends implicitly on $\boldsymbol{x}$. Thus, if the Implicit Function Theorem conditions hold, there is a function $\boldsymbol{a}_i$ of $\boldsymbol{x}$, such that $\boldsymbol{a}_i^* = \boldsymbol{a}_i(\boldsymbol{x})$ and we can define $h_i ( \boldsymbol{x} ) \triangleq g( \boldsymbol{a}_i ( \boldsymbol{x} ) , \boldsymbol{x} )$. Then, the \ac{EPF} $P ( \boldsymbol{x} )$ introduced in \cite{Watson1981} is defined in a neighborhood of $\boldsymbol{x}$ as:

\begin{equation}
P ( \boldsymbol{x} ) \triangleq f ( \boldsymbol{x}  )+\nu \sum_{i \in I ( \boldsymbol{x} )} h_i ( \boldsymbol{x} )_+
\label{eq:nine}
\end{equation}

\noindent where $\nu > 0$ and $h_i( \boldsymbol{x})_+ \triangleq \max  \{0,h_i ( \boldsymbol{x} ) \}$. We refer the interested reader to \cite{Watson1981} for more details about \acp{EPF}. Now, consider the \ac{EPF}-based optimization problem: %based on $P ( \boldsymbol{x} )$ and \pref{eq:general_SIP}:

\begin{equation}
	\begin{aligned}
		& \min_{\boldsymbol{x} \in \mathbb{R}^n }  P ( \boldsymbol{x} ) \\
		& \text{s.t.} \; l_u ( \boldsymbol{x} )=0, \forall u; \; r_v ( \boldsymbol{x} )\leq0, \forall v. 
	\end{aligned}
	\label{eq:general_EPF}
\end{equation}

\noindent
In the next theorem, we state that the general \ac{SIP} \pref{eq:general_SIP} can be solved by solving the \ac{EPF}-based \pref{eq:general_EPF}.

\begin{theorem}
\label{thm:equivalence_SIP_EPF}
Let $f$, $g$, $l_u$'s and $r_v$'s in \pref{eq:general_SIP} and \pref{eq:general_EPF} be continuously differentiable, and let the EPF function $P(\boldsymbol{x})$ be defined as in $\eref{eq:nine}$. Suppose for fixed $\boldsymbol{x}$, there are finite number of $\boldsymbol{a}_s \in \boldsymbol{\Omega}(\boldsymbol{a})$, $s = 1,\cdots, S$, such that $g(\boldsymbol{a}_s, \boldsymbol{x}) = 0$. If $\hat{\boldsymbol{x}}$ is in the feasible region of \pref{eq:general_SIP} and solves \pref{eq:general_EPF}, then $\hat{\boldsymbol{x}}$ is a \ac{KKT} point of \pref{eq:general_SIP}.
\end{theorem}

The proof of the above theorem is given in \sref{sec:Proof_to_Equivalence_SIP_EPF} of the Appendix. The \ac{EPF} optimization provides a way to solve the proposed general \ac{SIP}-type \pref{eq:general_SIP}; in addition, it also avoids involving the \ac{SIP} constraint. Obviously, since our framework of \pref{eq:equivalent_SIP} is a general SIP problem, we can solve its equivalent EPF-based problem instead. In the next section we introduce a simple and easy to implement algorithm to solve the \ac{EPF} optimization problem for our framework.

%%%%%%%%%%%%%%%%%%%%%%%%%%%%%%%%%%%%%%%%%%%%%%%%%%%%%%%%%%%%%%%%%%%%%%%%%%%%%%%%
%%%%%%%%%%%%%%%%%%%%%%%%%%%%%%%%%%%%%%%%%%%%%%%%%%%%%%%%%%%%%%%%%%%%%%%%%%%%%%%%
%%%%%%%%%%%%%%%%%%%%%%%%%%%%%%%%%%%%%%%%%%%%%%%%%%%%%%%%%%%%%%%%%%%%%%%%%%%%%%%%
\section{Algorithm}
\label{sec:Algorithm}

In this section, we focus on solving the EPF-based \pref{eq:general_EPF} for our framework. Consider \pref{eq:general_EPF}, for which we define $\boldsymbol{x} \triangleq  [ \omega, \boldsymbol{\theta}' ]'$, $\boldsymbol{a} \triangleq [ \boldsymbol{\alpha}^{1'}, \cdots, \boldsymbol{\alpha}^{T'} ]'$, $f ( \boldsymbol{x} ) \triangleq \omega$, $g ( \boldsymbol{a}, \boldsymbol{x} ) \triangleq \sum_{t=1}^T \bar{g}(\boldsymbol{\alpha}^t, \sum_{m=1}^M \theta_m^t \boldsymbol{K}_m^t) - \omega$, and let the constraints $l_u$'s and $r_v$'s define $\boldsymbol{\Psi} ( \boldsymbol{\theta} )$. In order to solve \pref{eq:general_EPF}, a descent algorithm is suggested in \cite{Watson1981}. During iteration $k$ and given $\boldsymbol{x}_k$, $E ( \boldsymbol{x}_k )$ is calculated, a descent direction $\boldsymbol{d}_k$ is found, and the variables are updated as $\boldsymbol{x}_{k+1} = \boldsymbol{x}_k + \epsilon \boldsymbol{d}_k$. We remind the reader that the set $E ( \boldsymbol{x} )$ contains $\boldsymbol{a}_i^*$'s, which satisfy $g(\boldsymbol{a}_i^*, \boldsymbol{x}) \geq -\eta$. In \cite{Watson1981} it is proven that, for sufficiently large $\nu$ and sufficiently small step length $\epsilon > 0$, the limit point of $\{ \boldsymbol{x}_k \}$ is a \ac{KKT} point of \pref{eq:general_SIP}, if the sequence $\{ \boldsymbol{d}_k \}$ is bounded and the sequence $\{ \boldsymbol{x}_k \}$ remains in a bounded region, in which $f$ and $g$ are doubly-differentiable functions with bounded second-order derivatives. In the next theorem, we show how to find the descent direction $\boldsymbol{d}_k$ for our framework given $\boldsymbol{x}_k$ at the $k$-th iteration.

\begin{theorem}
\label{thm:equivalent_min}
Let $\bar{g}(\boldsymbol{\alpha}, \boldsymbol{K})$ be affine with respect to the individual entries of $\boldsymbol{K}$. Suppose it has a finite maximum and a finite number of local maxima with respect to $\boldsymbol{\alpha}$ in the feasible set $\boldsymbol{\Omega}(\boldsymbol{\alpha})$. Let $l_u$'s and $r_v$'s be continuous and differentiable. Let $E(\boldsymbol{x})$ and $I(\boldsymbol{x})$ be as defined in \sref{sec:ExactPenaltyFunctionMethod}, and $\boldsymbol{a}_{i, k} \triangleq  [ \boldsymbol{\alpha}_{i, k}^{1'}, \cdots, \boldsymbol{\alpha}_{i, k}^{T'} ]' \in E( \boldsymbol{x}_k )$, $i \in I ( \boldsymbol{x}_k  ) = \{1,\cdots, N(\boldsymbol{x}_k) \}$. Consider the following $N(\boldsymbol{x}_k)$ problems, $\forall \; i \in I( \boldsymbol{x}_k )$:

\begin{equation}
\begin{aligned}
& \min_{\boldsymbol{\theta}}  \sum_{t=1}^T \bar{g}(\boldsymbol{\alpha}_{i, k}^t, \sum_{m=1}^M \theta_m^t \boldsymbol{K}_m^t) \\
& \text{s.t.} \; l_u ( \boldsymbol{x} )=0, \forall u; \; r_v ( \boldsymbol{x} )\leq0, \forall v. 
\end{aligned}
\label{eq:equivalent_min}
\end{equation}

\noindent Let $\hat{\boldsymbol{\theta}}_i$, consisting of all elements $\hat{\theta}_{m,i}^t$'s, be the solution to the $i$-th problem. Also, let  $i_0 = \arg\min_{i \in I ( \boldsymbol{x}_k  )}$ $ \max_{j \in I ( \boldsymbol{x}_k )} \{  \sum_{t=1}^T\bar{g} ( \boldsymbol{\alpha}_{j,k}^t, \sum_{m=1}^M \hat{\theta}_{m,i}^t\boldsymbol{K}_m^t ) \}$ and $\hat{\omega} $ $  = \min_{i \in I ( \boldsymbol{x}_k )}\max_{j \in I ( \boldsymbol{x}_k )} \{ \sum_{t=1}^T \bar{g} ( \boldsymbol{\alpha}_{j,k}^t, \sum_{m=1}^M \hat{\theta}_{m,i}^t\boldsymbol{K}_m^t \}$. Finally, let $\hat{\boldsymbol{x}}= [ \hat{\omega}, \hat{\boldsymbol{\theta}}_{i_0}' ]'$. Then, $\boldsymbol{d}_k = \hat{\boldsymbol{x}} - \boldsymbol{x}_k$ is a descent direction for the EPF-based \pref{eq:general_EPF} when $\nu > 1$.

\end{theorem}

\noindent 
The proof of the above theorem is provided in \sref{sec:proof_to_equivalent_min} of the appendix. Note that, if $\bar{g}$ is concave with respect to $\boldsymbol{\alpha}$, then $N(\boldsymbol{x}_k) = 1$. Thus, we have the following corollary.

\begin{corollary}
Let $\bar{g}(\boldsymbol{\alpha}, \boldsymbol{K})$ be affine with respect to the individual entries of $\boldsymbol{K}$ and concave with respect to $\boldsymbol{\alpha}$.  Let the $l_u$'s and $r_v$'s be continuous and differentiable. Also, let $E(\boldsymbol{x})$ and $I(\boldsymbol{x})$ be defined as in \sref{sec:ExactPenaltyFunctionMethod} and $\boldsymbol{a}_{k} \triangleq [ \boldsymbol{\alpha}_{k}^{1'}, \cdots, \boldsymbol{\alpha}_{k}^{T'}   ]' \in E ( \boldsymbol{x}_k )$. Consider the problem

\begin{equation}
\begin{aligned}
& \min_{\boldsymbol{\theta}}  \sum_{t=1}^T \bar{g}(\boldsymbol{\alpha}_k^t, \sum_{m=1}^M \theta_m^t \boldsymbol{K}_m^t) \\
& \text{s.t.} \; l_u ( \boldsymbol{x} )=0, \forall u; \; r_v ( \boldsymbol{x} )\leq0, \forall v. 
\end{aligned}
\label{eq:equivalent_min_concave}
\end{equation}

\noindent with solution $\hat{\boldsymbol{\theta}}$. Let $\hat{\omega} \triangleq \sum_{t=1}^T \bar{g}(\boldsymbol{\alpha}_k^t, \sum_{m=1}^M \hat{\theta}_m^t \boldsymbol{K}_m^t)$ and $\hat{\boldsymbol{x}} \triangleq [ \hat{\omega}, \hat{\boldsymbol{\theta}}' ]'$. Then, $\boldsymbol{d}_k = \hat{\boldsymbol{x}} - \boldsymbol{x}_k$ is a descent direction for the EPF-based \pref{eq:general_EPF} when $\nu > 1$.
\end{corollary}

Based on the above discussion, we provide \aref{alg:Algorithm}. Note that we focus on the case, where $ \bar{g}$ is concave, since it is the most common one in the context of kernel machines.

%\begin{algorithm}
%\caption{Algorithm for solving \pref{eq:general_EPF}}

\begin{algorithm}
\caption{Algorithm for solving \pref{eq:general_EPF}}
\label{alg:Algorithm}
\begin{algorithmic}
\STATE Choose $M$ kernel functions. Calculate the kernel matrices $\boldsymbol{K}_m^t$ for the $T$ tasks and the $M$ kernels. Choose $\bar{g}$ and $\boldsymbol{\Psi} \left ( \theta \right )$ based on the characteristics of the problem at hand. Randomly initialize $\boldsymbol{\theta}_{0} \succeq \boldsymbol{0}$. Initialize $\eta$ and $\epsilon_0$ to small positive values. $k \gets 0$.
\WHILE{not converged}
\STATE $\boldsymbol{a}_{k} \gets \arg\max_{\boldsymbol{a}\in\boldsymbol{\Omega ( \boldsymbol{a} )}} \sum_{t} \bar{g}(\boldsymbol{\alpha}^t, \sum_{m} \theta_{m, k}^t \boldsymbol{K}_m^t)$ 
\STATE $\hat{\boldsymbol{\theta}}_{k} \gets $ solve \pref{eq:equivalent_min_concave} given $\boldsymbol{a}_{k}$
\STATE $\hat{\omega}_k \gets \sum_{t} \bar{g}(\boldsymbol{\alpha}_k^t, \sum_{m} \hat{\theta}_{m, k}^t \boldsymbol{K}_m^t)$
\STATE $\boldsymbol{\theta}_{k+1} \gets \boldsymbol{\theta}_{k} + \epsilon_k ( \hat{\boldsymbol{\theta}}_{k} - \boldsymbol{\theta}_{k} )$
\STATE $\omega_{k+1} \gets \omega_k + \epsilon_k (\hat{\omega}_k - \omega_k)$
\STATE $k \gets k+1$
\ENDWHILE
\end{algorithmic}
\end{algorithm}

There are several ways to choose the step length $\epsilon_k$ in each step $k$. As suggested in \cite{Watson1981}, one possibility is to choose $\epsilon_k$ as the largest element of the set $ \{ 1, \beta, \beta^2, \cdots \}$, for some $\beta$, $0 < \beta < 1$, such that 

\begin{equation}
[P(\boldsymbol{x}_k + \epsilon_k \boldsymbol{d}_k) - P(\boldsymbol{x}_k)] / \epsilon_k G_k \geq \sigma
\label{eq:step_length}
\end{equation}

\noindent where $P$ is the EPF given in \pref{eq:general_EPF}, $\sigma$ is some constant that satisfies $0 < \sigma < 1$, and $G_k$ is the directional derivative of $P$ with respect to $\boldsymbol{x}$ at the $k$-th step. It is not difficult to show that 

\begin{equation}
G_k = f(\hat{\boldsymbol{x}}_k) + \nu \; g  ( \boldsymbol{a}_{k}, \hat{\boldsymbol{x}}_k )_+ -f(\boldsymbol{x}_{k}) - \nu \; g ( \boldsymbol{a}_{k}, \boldsymbol{x}_{k} )_+
\label{eq:directional_derivative}
\end{equation}

\noindent where $\hat{\boldsymbol{x}}_k = [ \hat{\omega}_k, \hat{\boldsymbol{\theta}}'_k ]'$ and $\boldsymbol{x}_k = [ \omega_k, \boldsymbol{\theta}'_k ]$. For details of $G_k$'s calculation, when $\bar{g}$ is not concave with respect to $\boldsymbol{\alpha}$, we refer the reader to \sref{sec:proof_to_equivalent_min} of the Appendix.

%%%%%%%%%%%%%%%%%%%%%%%%%%%%%%%%%%%%%%%%%%%%%%%%%%%%%%%%%%%%%%%%
\subsection{Analysis}
\label{sec:Analysis}

The advantages of our algorithm are multifold. First, the framework hinges on relatively few mild constraints that are typically met in practice. Specifically, we only assumed that $\bar{g}$ is continuous and doubly differentiable with respect to $\boldsymbol{\alpha}$ with bounded second-order derivatives and finite number of local maxima, whose value are also finite, and $\bar{g}$ is affine with respect to the elements of $\boldsymbol{K}$. There is no need for $\bar{g}$ to be concave with respect to $\boldsymbol{\alpha}$. Note also that all these constraints are met for the $4$ examples given in Problems \ref{eq:gBarSVM} through \ref{eq:gBar1CSVM}. Therefore, the framework and its associated algorithm may enjoy wide applicability.

Secondly, the maximization problem with respect to $\boldsymbol{a}$ in the first step of the algorithm can be separated into $T$ independent problems under the commonly-encountered setting, where the feasible regions of the $\boldsymbol{\alpha}^t$'s are mutually independent, such as in the formulations of \cite{Tang2009} and \cite{Rakotomamonjy2011}. Furthermore, in most situations, solving each of the $T$ problems can be addressed via readily-available efficient optimizers; for example, in the case of \ac{SVM} problems, one could use \emph{LIBSVM} \cite{CC01a}. In some other cases, a closed-form solution could be used instead, whenever available, such as in the case of \ac{KRR}. 

Thirdly, the minimization problem (\ref{eq:equivalent_min_concave}) is easy to solve. Since $\bar{g}$ is affine with respect to the individual entries of the kernel matrix, it is also affine in the entries of $\boldsymbol{\theta}$, which leads to the following optimization problem:

\begin{equation}
	\begin{aligned}
	& \min_{\boldsymbol{\theta}}  \boldsymbol{c}'\boldsymbol{\theta} \\
	& \text{s.t.} \; l_u ( \boldsymbol{x} )=0, \forall u; \; r_v ( \boldsymbol{x} )\leq0, \forall v. 
	\end{aligned}
\label{pr:optimize_theta}
\end{equation}

\noindent where $\boldsymbol{c}$ is a coefficient vector. For many practical models, the feasible regions of \pref{pr:optimize_theta} are such that a closed-form solution can be found. For example, in the case of an $L_p$-norm constrained (single-task) \ac{MKL} model, the feasible region is defined by the constraints $ \| \boldsymbol{\theta} \|_p \leq 1, \boldsymbol{\theta}\succeq \boldsymbol{0}$, for which a closed-form solution can be found. Another example is the setting considered in \cite{Tang2009}, which is of the form of \pref{eq:equivalent_min_concave} and becomes
%Another example is Tang's method \cite{Tang2009} (see \sref{sec:ProblemFormulation}). In applying our framework and algorithm to Tang's method, \pref{eq:equivalent_min_concave} becomes

\begin{equation}
	\begin{aligned}
	& \min_{\boldsymbol{\theta}}  \boldsymbol{c}'\boldsymbol{\theta} \\
	& \text{s.t.} \; \theta_m^t=\zeta_m + \gamma_m^t , \zeta_m \geq 0,\; \gamma_m^t \geq 0, \forall m, t; \\
	& \; \; \; \; \;\sum_{m=1}^M \theta_m^t = 1,\; \forall t; \; \sum_{m=1}^M \sum_{t=1}^T \gamma_m^t \leq \beta.
	\end{aligned}
\label{pr:tang_theta}
\end{equation}

\noindent Regarding the just-stated problem, a block coordinate descent algorithm can be employed to optimize $\boldsymbol{\zeta} = [\zeta_1,\cdots,\zeta_M]$ and $\boldsymbol{\gamma} = [{\boldsymbol{\gamma}^1}',\cdots,{\boldsymbol{\gamma}^T}']'$. Each sub-problem features a linear objective function with an $L_1$-norm constraint, and, hence, a closed-form solution can be found. Obviously, doing so is much simpler than relying on cutting-plane algorithms employed by the authors. Finally, a closed-form solution can be obtained for an $L_p-L_q$-norm constraint as well. \propref{prop:closed-form_solution_group_sparsity} offers closed-form solutions for problems featuring a linear objective with a $L_p$-norm or a $L_p-L_q$-norm constraint.

%\begin{proposition}
%\label{prop:closed-form_solution}
%Suppose $\boldsymbol{c}\in \mathbb{R}^n$ has at least one negative element, then the optimization problem

%\begin{equation}
%\begin{aligned}
%\min_{\boldsymbol{\theta}\in \mathbb{R}^n} & \hspace{1mm} \boldsymbol{c}^\prime\boldsymbol{\theta}\\
%\text{s.t.} \; & \boldsymbol{\theta} \succeq \boldsymbol{0}, \; \left \| \boldsymbol{\theta} \right \|_p \leq a.
%\end{aligned}
%\label{eq:lp_constraint_opt_prob}
%\end{equation}

%\noindent has closed-form solution 

%\begin{equation}
%\hat{\boldsymbol{\theta}} = a\frac{\left ( \tilde{\boldsymbol{c}}  \right )^{r}}{\left \| \tilde{\boldsymbol{c}} \right \|_{r+1}^{r}} 
%\label{eq:closed_form_lp_constraint}
%\end{equation}

%\noindent when $p > 1$, where $r = \frac{1}{p-1}$, $\tilde{\boldsymbol{c}}$  is the $n$-dimensional vector with element $\left [\max\left \{ -c_i, 0 \right \}  \right ] _{i=1}^n$, and $\left ( \tilde{\boldsymbol{c}}  \right )^{r}$ denotes element-wise exponentiation of vectors. When $p = 1$, the solution is $\hat{\boldsymbol{\theta}} = a\boldsymbol{e}_j$, where $\boldsymbol{e}_j$ is all-zero vector except the $j$-th element to be 1, and $j = \arg \min_{i}\left \{ c_1,\cdots,c_n \right \}$. If $\boldsymbol{c} \succeq \boldsymbol{0}$, then the solution is $\hat{\boldsymbol{\theta}} = \boldsymbol{0}$.

%\end{proposition}

\begin{proposition}
\label{prop:closed-form_solution_group_sparsity}
Let $\boldsymbol{c}  \triangleq [ c_1, \cdots, c_n ]' \in \mathbb{R}^n$ be the concatenation of $T$ vectors $\boldsymbol{c}^1 \in \mathbb{R}^{n_1}, \cdots, \boldsymbol{c}^T \in \mathbb{R}^{n_T}$ with $\sum_{t=1}^T n_t = n$. Suppose that, for $t = 1, \cdots, T$, each $\boldsymbol{c}^t$ has at least one negative element. Similarly, let $\boldsymbol{\theta} \triangleq [ \theta_1, \cdots, \theta_n ] \in \mathbb{R}^n $ be the concatenation of $\boldsymbol{\theta}^1 \in \mathbb{R}^{n_1}, \cdots, \boldsymbol{\theta}^T \in \mathbb{R}^{n_T}$. The optimization problem 

\begin{equation}
\begin{aligned}
\min_{\boldsymbol{\theta}\in \mathbb{R}^n} & \hspace{1mm} \boldsymbol{c}^\prime\boldsymbol{\theta}\\
\text{s.t.} \; & \boldsymbol{\theta} \succeq \boldsymbol{0}, \; (\sum_{t=1}^T \| \boldsymbol{\theta}^t \|_p^q )^{1/q} \leq a
\end{aligned}
\label{eq:group_sparsity_constraint_opt_prob}
\end{equation}

\noindent has closed-form solution 

\begin{equation}
\begin{cases}
\hat{\boldsymbol{\theta}}^t =  \frac{\sigma^t (\tilde{\boldsymbol{c}}^t )^r}{ \| \tilde{\boldsymbol{c}}^t \|_{r+1}^r},  \; \forall t, &   p > 1, q > 1\\ 
\hat{\boldsymbol{\theta}}^t =  \frac{ a(\tilde{\boldsymbol{c}}^t )^r}{ \| \tilde{\boldsymbol{c}}^t \|_{r+1}^r} [t = t_0] ,  \; \forall t, &   p > 1, q = 1\\
\hat{\boldsymbol{\theta}} = \frac{ a(\boldsymbol{c}_*)^s}{\| \boldsymbol{c}_* \|_{s+1}^s}, &   p = 1, q > 1\\
\hat{\boldsymbol{\theta}} = a\boldsymbol{e}_j,  &p = 1, q = 1
\end{cases}
\label{eq:closed_form_group_sparsity}
\end{equation}

\noindent where $\sigma^t \triangleq \frac{a \| \tilde{\boldsymbol{c}}^t \|_{r+1}^s}{ (\sum_{t=1}^T  \| \tilde{\boldsymbol{c}}^t \|_{r+1}^{s+1} )^{1/q} } $, $[t = t_0] = 1$, if $t = t_0$, and equals $0$, if otherwise, $t_0 \triangleq \arg\min_{t} $ $ \{ \| \tilde{\boldsymbol{c}}^t \|_{r+1} \}$, $r \triangleq \frac{1}{p-1}$, $s \triangleq \frac{1}{q-1}$, $\tilde{\boldsymbol{c}}^t$  is a vector with elements $[\max \{  -c_m^t, 0  \} ] _{m=1}^M$ and $( \tilde{\boldsymbol{c}}^t )^{r}$ denotes element-wise exponentiation of vectors. Also, $\boldsymbol{c}^*$ represents the concatenation of $\boldsymbol{c}_*^t \triangleq \boldsymbol{c}^t \circ \boldsymbol{e}^t_{i_t}, t = 1,\cdots, T$, where $\circ$ is element-wise multiplication of vectors and $\boldsymbol{e}_{i_t}^t \in \mathbb{R}^{n_t}$ is a vector, whose $i_t$-th entry is $1$, while the remaining are $0$, and $i_t \triangleq \arg \min_{i} \{ c^t_i \}$.  Moreover, $\boldsymbol{e}_j$ is an all-$0$ vector except that its $j$-th entry equals $1$, where $j \triangleq \arg \min_{i} \{ c_i \}$. Finally, if $\;\exists \; t$ such that $\boldsymbol{c}^t \succeq \boldsymbol{0}$, then $\hat{\boldsymbol{\theta}}^t = \boldsymbol{0}$.  
\end{proposition}

The proof of the above proposition is given in \sref{sec:proof_to_closed-form_solution_group_sparsity} of the Appendix. It is worth mentioning that, based on \propref{prop:closed-form_solution_group_sparsity}, true within-task sparsity is only achieved, when $p=1$. Similarly, true sparsity across tasks is only obtained, when $q=1$. Therefore, depending on the desired type of sparsity, different parameter settings should be applied. For example, if within-task sparsity is desired, an $L_1-L_2$ norm with $p=1$ and $q=2$ can be utilized, as discussed in \cite{Friedman2010}.

For a non-concave $\bar{g}$, but one that has a finite number of local maxima with respect to $\boldsymbol{\alpha}$ given $\boldsymbol{x}_k$, based on \thmref{thm:equivalent_min}, we only need to solve $N(\boldsymbol{x}_k)$ problems that are similar to \pref{eq:equivalent_min_concave}, which in most cases have closed-form solutions, as stated earlier. 

Based on the above analysis, it is not difficult to see that, when a closed-form solution can be found in the second step, the complexity of our algorithm in each iteration is dominated by the complexity of the solver of the first step, such as the \ac{SVM}, \ac{SVDD} solver, etc. The time complexity of \emph{LIBSVM} for solving a \ac{SVM} or \ac{SVDD} problem is given in \cite{CC01a}.  A \ac{KRR}-based model, as stated earlier, can be solved in constant time. Therefore, if the second step has a closed-form solution, then the complexity of each iteration for such model is $O(1)$. Unsurprisingly, we observed in practice that our algorithm is usually slower than the special algorithms that are tailored to each specific problem type. For example, for single-task MKL with an $L_p$-norm constraint as discussed in \cite{Xu2010}, a block coordinate descent algorithm is used, which is equivalent to our algorithm with step length equal to 1, when our algorithm is adapted to solve this model. Thus, since our algorithm initializes the step length to be a small value, it is not a surprise that our algorithm is slower. However, if the step length of our algorithm is initialized to $1$, both of the two methods are exactly the same. Additionally, for solving the model that is proposed in \cite{Tang2009}, when $500$ samples from the USPS data set are used for training, our algorithm takes on average $9$ seconds to train the model, while the algorithm proposed in \cite{Tang2009} takes $7$ seconds. As expected, the advantages of our algorithm, namely its simplicity in terms of implementation and its generality are offset by a somewhat reduced computational efficiency, when compared to highly specialized algorithms that are designed to solve very specific problems.

\section{The Partially Shared Common Space Model}
\label{sec:PSCS_model}

To demonstrate the flexibility of our framework and the capability of the associated algorithm, in this section we propose a \acf{PSCS} model for \ac{MT-MKL} as a novel concrete instance of our framework. As stated earlier, in \ac{MTL}, several tasks sharing a common feature representation are trained simultaneously. For \ac{MT-MKL}, it is a natural choice to let all tasks share a common kernel function by letting $\theta_m^t = \zeta_m, \forall \; t$ (see \sref{sec:ProblemFormulation}). However, in practice, there may be some problems, for which sharing a common space is not the optimal choice. For example, an \ac{MTL} problem may include a few complex tasks, but also some much simpler ones. In this situation, it may be difficult to find a common feature mapping so that all tasks perform well. Therefore, it is meaningful to let complex tasks to use their own task-specific space, while allowing the remaining tasks to share a common space. Motivated by this consideration, we introduce the \ac{PSCS} problem formulation shown below:

\begin{equation}
\begin{aligned}
& \min_{\boldsymbol{\theta}, \boldsymbol{\zeta}, \boldsymbol{\gamma}} \max_{\boldsymbol{a} \in \boldsymbol{\Omega} ( \boldsymbol{a} )} \sum_{t=1}^T \bar{g}(\boldsymbol{\alpha}^t, \sum_{m=1}^M \theta_m^t \boldsymbol{K}_m^t) \\ 
& \text{s.t.} \; \theta_m^t = \zeta_m + \gamma_m^t, \; \forall m,\; t; \\
& \boldsymbol{\zeta} \succeq \boldsymbol{0}, \; \| \boldsymbol{\zeta} \|_p \leq 1, \; p \geq 1; \\
& \boldsymbol{\gamma} \succeq \boldsymbol{0}, \;  (\sum_{t=1}^T \| \boldsymbol{\gamma}^t \|_p^q  )^{1/q} \leq 1, \; p \geq 1, \; q \geq 1.
\end{aligned}
\label{eq:PSCS_model}
\end{equation}

The feasible region of $\boldsymbol{\theta}$ is chosen to meet our objectives: the $L_p$-norm constraint controls the sparsity of the common-space coefficients $\boldsymbol{\zeta}$, while the $L_p-L_q$-norm constraint controls the within-group and group-wise sparsity. Choosing $q=1$ induces sparcity on $\boldsymbol{\gamma}$ (group-wise sparsity), which will force most tasks to share a common feature space. 

Due to the complicated nature of the constraints, it is far from straightforward to devise a simple, tailored algorithm to solve the PSCS model. However, \aref{alg:Algorithm} can be readily applied. In each iteration, $T$ kernel machines are first trained. Next, the minimization with respect to $\boldsymbol{\theta}$ can be accomplished via block coordinate descent to optimize $\boldsymbol{\zeta}$ as a block and $\boldsymbol{\gamma}$ as another block. In addition, the closed-form solution for $\boldsymbol{\theta}$ and $\boldsymbol{\gamma}$ can be easily determined by \propref{prop:closed-form_solution_group_sparsity}.

%it is not difficult to show that closed-form solutions can be found for both of the two group of variables:

%\begin{equation}
%\boldsymbol{\zeta} = \frac{\left (\sum_{t=1}^T \boldsymbol{G}^t  \right )^r}{\left \|\sum_{t=1}^T \boldsymbol{G}^t  \right \|_{r+1}^r},   \;\;\; \boldsymbol{\gamma}^t = \frac{\sigma^t\left (\boldsymbol{G}^t  \right )^r}{\left \| \boldsymbol{G}^t \right \|_{r+1}^r},   \;\;\;  \forall t = 1,\cdots,T
%\label{eq:PSCS_solutions}
%\end{equation}

%\noindent where $\sigma^t = \frac{\left \| \boldsymbol{G}^t \right \|_{r+1}^s}{\left (\sum_{t=1}^T \left \| \boldsymbol{G}^t \right \|_{r+1}^{s+1}  \right )^{1/q} }$ $\boldsymbol{G}^t = \left [ G_1^t,\cdots,G_M^t \right ]'$, $G_m^t = \boldsymbol{\alpha}^{t'} \boldsymbol{Y}^t  \boldsymbol{K}_m^t \boldsymbol{Y}^t \boldsymbol{\alpha}^t$, $r = \frac{1}{p-1}$ and $s = \frac{1}{q-1}$. 

%%%%%%%%%%%%%%%%%%%%%%%%%%%%%%%%%%%%%%%%%%%%%%%%%%%%%%%%%%%%%%%%%%%%%%%%%%%%%%%%
%%%%%%%%%%%%%%%%%%%%%%%%%%%%%%%%%%%%%%%%%%%%%%%%%%%%%%%%%%%%%%%%%%%%%%%%%%%%%%%%
%%%%%%%%%%%%%%%%%%%%%%%%%%%%%%%%%%%%%%%%%%%%%%%%%%%%%%%%%%%%%%%%%%%%%%%%%%%%%%%%
\section{Experimental Results}
\label{sec:ExperimentalResults}

In the following subsections, we experimentally evaluate our \ac{PSCS} model on classification tasks. To apply our framework for classification tasks, the objective function $\bar{g}$ is specified as the dual-domain objective function for \ac{SVM} training:

\begin{equation}
\sum_{t=1}^T \bar{g}(\boldsymbol{\alpha}^t, \sum_{m=1}^M \theta_m^t\boldsymbol{K}_m^t)
= \sum_{t=1}^T ({\boldsymbol{\alpha}^t}' \boldsymbol{1} - {\boldsymbol{\alpha}^t}' \boldsymbol{Y}^t (\sum_{m=1}^M \theta_m^t \boldsymbol{K}_m^t) \boldsymbol{Y}^t \boldsymbol{\alpha}^t )
\label{eq:SVMObjective}
\end{equation}

\noindent
Note that all kernel functions in our experiments are used in their normalized form as %via $k(\boldsymbol{x}, \boldsymbol{y}) \leftarrow 
$\frac{k(\boldsymbol{x}, \boldsymbol{y})}{\sqrt{k(\boldsymbol{x}, \boldsymbol{x}) k(\boldsymbol{y}, \boldsymbol{y})}}$. 
%%%%%%%%%%%%%%%%%%%%%%%%%%%%%%%%%%%%%%%%%%%%%%%%%%%%%%%%%%%%%%%%%%%%%%%%%%%%%%%%
%%%%%%%%%%%%%%%%%%%%%%%%%%%%%%%%%%%%%%%%%%%%%%%%%%%%%%%%%%%%%%%%%%%%%%%%%%%%%%%%
\subsection{A qualitative case study}
\label{subsec:IrisDataSet}

To qualitatively illustrate the potential of our framework, in this subsection we apply our approach to the well-known \textit{Iris Flower} classification problem, which we will recast as a \ac{MTL} problem. The associated data set includes $150$ patterns, each of which comes from one of three Iris flower classes: \emph{Setosa}, \emph{Versicolour} and \emph{Virginica} (respectively, class $1$, $2$ and $3$). Each of these $3$ classes is represented by $50$ samples and every sample has $4$ attributes corresponding to the width and length of the flower's sepal and pedal. We chose only two attributes, namely the sepal width and length, to form a $2$-dimensional data set, such that the distribution of patterns and the resulting decision boundaries can be visualized. Note that each attribute is normalized to $[0, 1]$. We split the three-class problem into three binary classification tasks by employing a one-against-one strategy. Specifically, these are: task $1$ (class $1$ vs. class $2$), task $2$ (class $1$ vs. class $3$), task $3$ (class $2$ vs. class $3$). The data sets of task $1$ and $2$ are linearly separable, so it is desirable to obtain classifiers which will produce linear or almost-linear decision boundaries. On the other hand, the data set of task $3$ is not linearly separable. Intuitively, one would expect a reasonable solution to have tasks $1$ and $2$ share a common feature space, while allowing task $3$ to be mapped into an alternative, task-specific feature space.

\begin{figure}[htpb]
	\centering
	\includegraphics[width=3.2in]{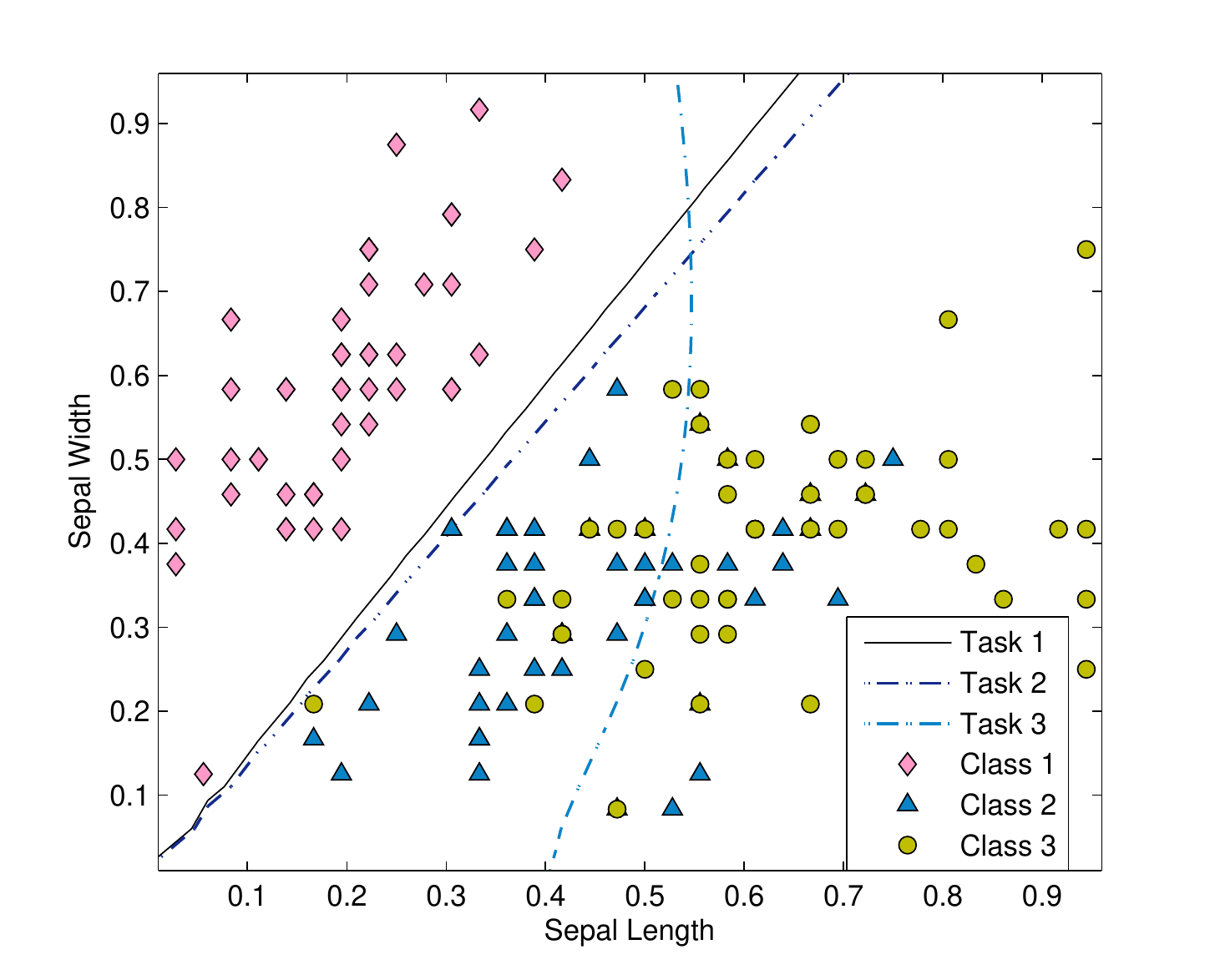}
	\caption{Decision boundaries of \textit{Iris} classification multi-task problem.}	         
	\label{fig:Iris}
\end{figure}

For the purpose of this experiment, we employed Linear kernels, Polynomial kernels of degree $2$, and Gaussian kernels using a spread parameter value of $5$. All $150$ patterns were used for training. The $p$ parameter was set to $2$ for non-sparse kernel combination, because we want to see how different tasks affect the weights of each kernel function. Additionally, $q$ is set to $1$, because we want to achieve inter-task sparsity for $\boldsymbol{\gamma}$, \ie\ some tasks share a common space specified by $\boldsymbol{\zeta}$. The experimental results are shown in \tref{tab:Iris} and \fref{fig:Iris}.

\begin{table}[htpb]
\begin{center}
\caption{Learned coefficients for the \textit{Iris} classification multi-task problem} 
\label{tab:Iris}
\begin{tabular}{ccccc}
\toprule
Kernel & $\boldsymbol{\zeta}$ & $\boldsymbol{\gamma}^1$ & $\boldsymbol{\gamma}^2$ & $\boldsymbol{\gamma}^3$ \\
\midrule
Linear & 0.1828 & 0 & 0 & 0.0295 \\
Polynomial &0.9421 & 0 & 0 & 0.1976 \\
Gaussian & 0.2812 & 0 & 0 & 0.9333 \\
\bottomrule
\end{tabular}
\end{center}
\end{table}

It can be seen in \tref{tab:Iris} that all three elements in $\boldsymbol{\gamma}^1$ and $\boldsymbol{\gamma}^2$ are zero. This means that task $1$ and $2$ share a common space, which is specified by $\boldsymbol{\zeta} = [0.1828, 0.9421, 0.2812]$. Obviously, in the common space, the combination of Linear and Polynomial kernels has a large weight, while the Gaussian kernel plays an insignificant role. The reason behind this is that the data for tasks $1$ and $2$ are both linearly separable, so the \ac{SVM} can easily find effective boundaries in the original space and the feature space implied by the Polynomial kernel. Also, we can observe that the polynomial kernel has larger weight than the linear kernel. This is because the data in the feature space induced by the polynomial kernel offers a larger margin than in the case of the original space. %Therefore, the polynomial kernel induced feature space is more preferred, and thus it has larger weight. 
Note that, even though the polynomial kernel does not imply a linear feature map, the decision boundary is almost a straight line. This implies that the mapping corresponding to the polynomial kernel is almost linear in this particular case. Unlike tasks $1$ and $2$, we can see from the table that the task-specific feature space for task $3$ has a large weight corresponding to the Gaussian kernel. This is because the data for task $3$ are not linearly separable, and therefore it is difficult for the associated \ac{SVM} to find a good decision boundary in the original space and the feature space implied by the Polynomial kernel. A large weight corresponding to the Gaussian kernel implies that the classifier was able to find a better decision boundary in its associated infinite-dimensional feature space. The relevant decision boundaries can be seen in \fref{fig:Iris}. As expected, the decision boundaries for tasks $1$ and $2$ are almost linear, while task $3$ features a non-linear decision boundary.

%%%%%%%%%%%%%%%%%%%%%%%%%%%%%%%%%%%%%%%%%%%%%%%%%%%%%%%%%%%%%%%%%%%%%%%%%%%%%%%%
%%%%%%%%%%%%%%%%%%%%%%%%%%%%%%%%%%%%%%%%%%%%%%%%%%%%%%%%%%%%%%%%%%%%%%%%%%%%%%%%
\subsection{Quantitative Analysis on Benchmark Problems}
\label{sec:OtherBenchmarkProblems}

In this subsection we first evaluate our \ac{SVM}-based \ac{PSCS} method using $6$ benchmark multi-class data sets obtained from the UCI repository \cite{Frank2010}. Each associated recognition problem was cast as a multi-task classification problem by using the one-against-all approach. This is an effective method to test multi-task models; for example, refer to \cite{Jebara2011} and \cite{Zhong2012}. More specifically, we used the USPS Handwritten Digit (\textit{USPS}), MNIST Handwritten Digit (\textit{MNIST}), Wall-Following Robot Navigation (\textit{Robot}), Statlog Shuttle (\textit{Shuttle}), Statlog Vehicle Silhouettes (\textit{Vehicle}), and Letter Recognition (\textit{Letter}) data sets.  For each data set, each class is represented by an equal number of samples. An exception is the original \textit{Shuttle} data set that has seven classes, four of which are very poorly represented; for this data set we only chose data from the other three classes. Finally, for the \textit{Letter} data set, we only chose the first $10$ classes to avoid handling a large number tasks.

%For all datasets, we randomly sample a subset of the data due to their large size. The number of samples used in the experiments and other characteristics of the datasets can be found in \tref{Table:Data}.

%\begin{table}[htpb]
%\begin{center}
%\caption{Characteristics of the Benchmark Datasets} % How to increase the distance between the title and the table????
%\label{Table:Data}
%\begin{tabular}{lcccccc}
%\toprule
% & USPS & MNIST & Robot & Shuttle & Vehicle & Letter \\
%\midrule
%\# Sample & 1000 & 1000 & 1000 & 900 & 846 & 1000 \\
%\# Classes & 10 & 10 & 4 & 3 & 4 & 10 \\
%\# Features & 256 & 784 & 4 & 9 & 18 & 16\\
%\bottomrule
%\end{tabular}
%\end{center}
%\end{table}

For all experiments, the kernel and algorithm parameter settings were held fixed. Twenty experiments were conducted for each setting and the average correct classification accuracy was recorded. Linear, Polynomial, and Gaussian kernels with spread parameter values $\left \{ 2^0, 2^1, 2^2, 2^3, 2^4, 2^5, 2^6, 2^7 \right \}$ were used. Regarding $p$, we held it fixed to $1.1$, as we were not concerned with obtaining within-group sparsity. Cross-validation was employed to choose values for parameters $C$ and $q$. We allowed $C$ to vary over $\left \{ 1/27, 1/9, 1/3, 1, 3, 9, 27 \right \}$ and $q$ ranged from $1.0$ to $2.0$ with a step size of $0.1$. We did not consider the case, where $q > 2$, since these values would yield a non-sparse group-wise $\boldsymbol{\gamma}$ vector, which was not deemed desirable. 

The \ac{PSCS} model was compared to the CS and IS methods, which have been described in \sref{sec:ProblemFormulation}. The experimental settings for the CS and IS methods were exactly the same as the settings used for PSCS, except that there was no parameter $q$ to be tuned for them. We considered training set sizes of $2$\%, $5$\%, $20$\% and $50$\% of the original data set to study the effect of training set size on classification accuracy.  The rest of the data were split in half to form validation and test sets.

We report the experimental results (average classification accuracy of $20$ runs over randomly sampled training set) in \tref{tab:Result1}, where the best performance is highlighted in boldface. To test the statistical significance of the differences between the best performing method and the rest, we employed a t-test to compare mean accuracies using a significance level of $\alpha=0.05$. Furthermore, we use two superscripts for each \ac{PSCS}-related result to convey information about statistically significant differences in performance. The first sign refers to the comparison of \ac{PSCS} to CS and the second one to the comparison to IS. A `$+$' sign means that the \ac{PSCS} performance is significantly better, and a `$=$' sign indicates that there is no statistically significant difference.

\begin{table}[htpb]
\begin{center}
\caption{Experimental comparison of PSCS to the CS, IS methods on six benchmark data sets in terms of percent correct classification accuracy.}
\label{tab:Result1}
\begin{tabular}{l l l l l}
\toprule
Robot & 2\% & 5\% & 20\% & 50\%  \\
\midrule
PSCS & $\textbf{75.58}^{++}$ & $\textbf{82.21}^{++}$ & $\textbf{90.02}^{++}$ & $\textbf{92.27}^{==}$ \\
CS & 73.01 & 78.76 & 88.49 & 91.49 \\
IS & 73.09 & 78.91 & 87.86 & 91.76 \\
\midrule
Vehicle & 2\% & 5\% & 20\% & 50\% \\
\midrule
PSCS & $\textbf{51.25}^{==}$ & $\textbf{64.42}^{++}$ & $\textbf{76.05}^{=+}$ & $\textbf{82.35}^{++}$ \\
CS & 48.78 & 61.24 & 75.58 & 81.47 \\
IS & 48.62 & 61.76 & 75.04 & 80.62 \\
\midrule
USPS & 2\% & 5\% & 20\% & 50\% \\
\midrule
PSCS & $\textbf{52.78}^{++}$ & $\textbf{68.41}^{=+}$ & $87.03^{==}$ & $\textbf{93.34}^{==}$ \\
CS & 50.11 & 68.39 & \textbf{87.36} & 92.99 \\
IS & 46.52 & 63.95 & 86.91 & 92.78 \\
\midrule
Shuttle & 2\% & 5\% & 20\% & 50\% \\
\midrule
PSCS & $\textbf{93.63}^{++}$ & $\textbf{93.91}^{++}$ & $\textbf{94.90}^{++}$ & $\textbf{94.93}^{==}$ \\
CS & 91.80 & 91.84 & 94.20 & 94.90 \\
IS & 92.46 & 92.53 & 94.15 & 94.66 \\
\midrule
Letter & 2\% & 5\% & 20\% & 50\% \\
\midrule
PSCS & $50.19^{==}$ & $\textbf{67.43}^{++}$ & $82.65^{==}$ & $\textbf{89.23}^{==}$ \\
CS & 48.82 & 63.90 & 82.48 & 89.12 \\
IS & \textbf{51.35} & 64.21 & \textbf{82.99} & 89.16 \\
\midrule
MNIST & 2\% & 5\% & 20\% & 50\% \\
\midrule
PSCS & $55.15^{=+}$ & $\textbf{70.59}^{=+}$ & $84.38^{==}$ & $89.92^{==}$ \\
CS & \textbf{55.32} & 69.18 & \textbf{84.84}& 89.81 \\
IS & 51.29 & 66.59 & 84.34 & \textbf{90.13} \\
\bottomrule
\end{tabular}
\end{center}
\end{table}

It can be seen from \tref{tab:Result1} that, for small training set sizes ($2$\% and $5$\%), the \ac{PSCS} method is almost always better in terms of classification accuracy compared to the CS and IS approaches and that most of the differences are statistically significant. In other words, there are learning tasks that, in practice, may benefit from partially sharing a common feature space. Additionally, for all three methods considered, as the training set size increases to $20$\% or higher, their performance tends to be similar for most data sets; \ac{PSCS} seems to perform significantly better only for the \textit{Vehicle} data set. It appears that, when the training set is large enough, each task independently can be trained well, while sharing a common space among tasks helps little to further enhance performance.

In what follows we compare the \ac{PSCS} method to the CS and IS approaches on two widely-used multi-task classification data sets, namely the \emph{Letter} and \emph{Landmine} data sets. The \emph{Letter} data set\footnote{Available at: \url{http://multitask.cs.berkeley.edu/}} is a collection of handwritten words compiled by Rob Kassel of the MIT Spoken Language Systems Group. The associated \ac{MTL} problem involves 8 tasks, each of which is a binary classification problem. The 8 tasks are: `C' \vs\ `E', `G' \vs\ `Y', `M' \vs\ `N', `A' \vs\ `G', `I' \vs\ `J', `A' \vs\ `O', `F' \vs\ `T' and `H' \vs\ `N'. Each letter is represented by a $8 \times 16$ pixel image, hence by a $128$ dimensional feature vector. The goal for this problem is to correctly recognize the letters in each task. For this data set, our intention was to compare the performance of the three models on large data sizes. Therefore, we chose $1000$ samples from each class, so that each task had $2000$ training samples. Note that, due to insufficient data for classes `J', `H' and `F', our multi-task recognition problem considered only five of the eight tasks.

On the other hand, the \emph{Landmine} data set\footnote{Available at: \url{http://people.ee.duke.edu/~lcarin/LandmineData.zip}} consists of $29$ binary classification tasks. Each datum is a $9$-dimensional feature vector extracted from radar images that capture a single region, which may contain a landmine field. The $9$ features include four moment-based features, three correlation-based features, one energy ration feature and one spatial variance feature \cite{Xue2007}. Tasks $1-15$ correspond to regions with relatively high foliage, while the other $14$ tasks correspond to regions that are bare earth or desert. The tasks entail different amounts of data, varying from $30$ to $96$ samples. The goal is to classify regions as ones containing landmine fields or not. %The goal is to classify if the test data contains landmine or not in the specific region. 

\begin{table}[htpb]
\begin{center}
\caption{Experimental comparison of PSCS to the CS, IS methods on multi-task data sets in terms of percent correct classification accuracy.}
\label{tab:ResultMultiTask}
\begin{tabular}{l l l l l}
\toprule
Letter & 0.2\% & 0.5\% & 5\% & 50\%  \\
\midrule
PSCS & $\textbf{59.86}^{==}$ & $\textbf{67.02}^{++}$ & $\textbf{88.02}^{==}$ & $\textbf{94.29}^{==}$ \\
CS & 56.52 & 65.39 & 87.68 & 94.31 \\
IS & 56.41 & 63.85 & 87.80 & 94.45 \\
\midrule
Landmine & 20\% & 30\% & 40\% & 50\%  \\
\midrule
PSCS & $\textbf{69.89}^{++}$ & $\textbf{73.82}^{++}$ & $\textbf{75.12}^{=+}$ & $\textbf{77.15}^{=+}$ \\
CS & 67.24 & 71.62 & 75.04 & 76.96 \\
IS & 66.34 & 70.91 & 74.30 & 76.41 \\
\bottomrule
\end{tabular}
\end{center}
\end{table}

The experimental setting for these two problems were the same as previous experiments, except that we did not use $2$\% and $5$\% of the \emph{Landmine} data set for training, due to the small size of the data set. Instead, we started from $20$\% and increased the training set size in steps of $10$\%. Also, for the \emph{Letter} data set, in order to show results for a wider range of training set sizes, we chose $0.2$\%, $0.5$\%, $5$\% and $50$\% of the data set for training. 

The experimental results are displayed in \tref{tab:ResultMultiTask}, from which it can be observed that the experimental results are consistent with the ones obtained in the previous subsection. The \ac{PSCS} method can improve the classification accuracy significantly, when the training set size is relatively small, while in the case of large training sets, the improvements are not statistically significant. In the latter situation, all three methods perform similarly. 

It is also worth mentioning that, by using our proposed algorithm, all three models have the same asymptotic computation complexity. Obviously, in each iteration, all three models solve the same $T$ \ac{SVM} problems followed by solving \pref{eq:equivalent_min_concave}. Obviously, the former step differs for each method; \pref{eq:equivalent_min_concave} has a closed-form solution based on \propref{prop:closed-form_solution_group_sparsity} for each of the models. Therefore, the computational complexity per iteration is common among them and is dominated by the complexity of solving the $T$ \ac{SVM} problems. 

On balance, since \ac{PSCS} has no worse asymptotic runtime complexity, when compared to the IS and CS approaches, and since it offers a performance advantage in cases of sample scarcity, the use of \ac{PSCS} seems much preferable over the IS and CS formulations.

% reset all acronyms
\acresetall

%%%%%%%%%%%%%%%%%%%%%%%%%%%%%%%%%%%%%%%%%%%%%%%%%%%%%%%%%%%%%%%%%%%%%%%%%%%%%%%%
%%%%%%%%%%%%%%%%%%%%%%%%%%%%%%%%%%%%%%%%%%%%%%%%%%%%%%%%%%%%%%%%%%%%%%%%%%%%%%%%
%%%%%%%%%%%%%%%%%%%%%%%%%%%%%%%%%%%%%%%%%%%%%%%%%%%%%%%%%%%%%%%%%%%%%%%%%%%%%%%%
\section{Conclusions}
\label{sec:Conclusions}

In this paper, we proposed a \ac{MT-MKL} framework, which is formulated as a min-max problem and which subsumes a broad class of kernel-based learning problems. We showed that our formulation can be optimized by solving an \ac{EPF} optimization problem. Subsequently, we derived a simple algorithm to solve it, which, for some frequently-used learning tasks, is able to leverage from existing efficient solvers or from closed-form solutions. The availability of this algorithm eliminates the need of using existing, potentially much sophisticated, algorithms or devising algorithms that are specifically tailored to the problem at hand.

In order to illustrate the utility of this novel framework and associated algorithm, we devised the \ac{PSCS} \ac{MT-MKL} model as a special case of our framework, which allows some tasks to share a common feature space, while other tasks are allowed to be appropriately associated to their own task-specific feature spaces. Results obtained from experimenting with a collection of classification tasks demonstrated performance advantages of the \ac{PSCS} formulation, especially in the case, where the amount of training data is limited.

\section*{Acknowledgements}
\addcontentsline{toc}{section}{Acknowledgments}

C. Li acknowledges partial support from \ac{NSF} grant No. 0806931 and No. 0963146. Moreover, M. Georgiopoulos acknowledges partial support from \ac{NSF} grants No. 0525429, No. 0963146, No. 1200566 and No. 1161228. Finally, G. C. Anagnostopoulos acknowledges partial support from \ac{NSF} grant No. 0647018. Any opinions, findings, and conclusions or recommendations expressed in this material are those of the authors and do not necessarily reflect the views of the \ac{NSF}. Finally, the authors would like to thank the anonymous reviewers of this manuscript for their time and helpful comments.

% The plainrul style will printout the URL field of each bib entry
% and hyperref will create a clickable link.
\bibliographystyle{plainurl}
% Modify the bibliography file name as necessary.
\bibliography{IEEE-TNNLS2012paperA}

% Remove if you don't have appendix.

%\section*{Appendix}
%\label{sec:Appendix}
\appendix

\subsection{Proof to \thmref{thm:equivalence_SIP_EPF}}
\label{sec:Proof_to_Equivalence_SIP_EPF}

First, note that if $\hat{\boldsymbol{x}} $ solves \pref{eq:general_SIP}, then the $\hat{\boldsymbol{a}}_s$'s, which satisfy $g ( \hat{\boldsymbol{a}}_s, \hat{\boldsymbol{x}} ) = 0$ and $\hat{\boldsymbol{a}}_s \in \boldsymbol{\Omega} (\boldsymbol{a})$, are local maxima of $g ( \boldsymbol{a}, \hat{\boldsymbol{x}} )$. By assumption, there is a finite number of such $\hat{\boldsymbol{a}}_s$'s, say, $S$, which means that the problem has only $S$ active constraints. We can therefore provide the KKT necessary conditions for $\hat{\boldsymbol{x}}$ to be a solution of \pref{eq:general_SIP}:

\noindent
\textbf{KKT conditions for \pref{eq:general_SIP}} Let $\hat{\boldsymbol{x}}$ be a solution of \pref{eq:general_SIP}. Then, there exist $\hat{\lambda}_s$, $s=0,\cdots,S$; $\hat{\rho}_u$, $u=1,\cdots,U$; $\hat{\mu}_v$, $v=1,\cdots,V$, not all zero, such that

\begin{equation}
	\hat{\lambda}_0\phi_j ( \hat{\boldsymbol{x}} )+\sum_{s=1}^S \hat{\lambda}_s \psi_j ( \hat{\boldsymbol{a}}_s, \hat{\boldsymbol{x}} )+\sum_{u=1}^U \hat{\rho}_u \tau_{u,j}( \hat{\boldsymbol{x}} ) +\sum_{v=1}^V \hat{\mu}_v \omega_{v,j} ( \hat{\boldsymbol{x}} )=0, \; j=1,\cdots,n
	\label{eq:KKT_equality}
\end{equation}

\noindent where $\phi_j( \hat{\boldsymbol{x}})$, $\psi_j ( \hat{\boldsymbol{a}}_s, \hat{\boldsymbol{x}} )$, $\tau_{u,j} ( \hat{\boldsymbol{x}} )$, $\omega_{v,j} ( \hat{\boldsymbol{x}} )$  denote the quantities $\frac{\partial f ( \boldsymbol{x})}{\partial x_j} |_{\boldsymbol{x} = \hat{\boldsymbol{x}}}$, $ \frac{\partial g ( \hat{\boldsymbol{a}}_s, \boldsymbol{x} )}{\partial x_j}|_{\boldsymbol{x} = \hat{\boldsymbol{x}}}$, $ \frac{\partial l_u (\boldsymbol{x} )}{\partial x_j} |_{\boldsymbol{x} = \hat{\boldsymbol{x}}}$, $ \frac{\partial r_v (\boldsymbol{x} )}{\partial x_j}|_{\boldsymbol{x} = \hat{\boldsymbol{x}}}$ respectively.

This is a direct extension of the KKT conditions of the following standard SIP problem provided in \cite{Watson1981}:
 
\begin{equation}
\min_{\boldsymbol{x}\in \mathbb{R}^n } f(\boldsymbol{x}) \; \text{s.t.} \; g( \boldsymbol{a}, \boldsymbol{x} ) \leq 0, \; \forall \boldsymbol{a} \in \boldsymbol{\Omega}  ( \boldsymbol{a} ). 
\label{pr:standard_SIP}
\end{equation}

Similarly, we can state the KKT conditions for the EPF-based \pref{eq:general_EPF}:

\noindent
\textbf{KKT conditions for \pref{eq:general_EPF}} Let $\hat{\boldsymbol{x}}$ be a solution of \pref{eq:general_EPF}. Then, there exist $\hat{\lambda}_0'$ and $\hat{\lambda}_s', \; s \in I ( \hat{\boldsymbol{x}} ); \; \hat{\rho}_u', \; u=1,\cdots,U; \; \hat{\mu}_v', \; v=1,\cdots,V $, not all zero, such that

\begin{equation}
	\hat{\lambda}_0' \phi_j ( \hat{\boldsymbol{x}} )+\sum_{s \in I  ( \hat{\boldsymbol{x}} )} \hat{\lambda}_s' \psi_j ( \hat{\boldsymbol{a}}_s, \hat{\boldsymbol{x}} ) + \sum_{u=1}^U \hat{\rho}_u' \tau_{u,j} ( \hat{\boldsymbol{x}} )  + \sum_{v=1}^V \hat{\mu}_v' \omega_{v,j} ( \hat{\boldsymbol{x}} )=0, j=1,\cdots,n 
\label{eq:EPF_KKT_equation}
\end{equation}

\noindent In light of (\ref{eq:KKT_equality}) and (\ref{eq:EPF_KKT_equation}), we arrive at the conclusion.

%%%%%%%%%%%%%%%%%%%%%%%%%%%%%%%%%%%%%%%%%%%%%%%%%%%%%%%%%%%%%%%%%%%%%%%%%%%%%%%%%
\subsection{Proof to \thmref{thm:equivalent_min}}
\label{sec:proof_to_equivalent_min}

Before delving into the details, we provide a sketch of the proof. The proof itself includes two major stages. In the first stage, we prove that $\boldsymbol{d}_{k} = \hat{\boldsymbol{x}} - \boldsymbol{x}_{k}$ is a descent direction, if $\hat{\boldsymbol{x}}_{k}$ is the solution to the following problem:

\begin{equation}
\begin{aligned}
& \min_{\boldsymbol{x}} f ( \boldsymbol{x}  )+\nu \sum_{i \in I ( \boldsymbol{x}_{k} )} g ( \boldsymbol{a}_{i, k}, \boldsymbol{x} )_+ \\
& \text{s.t.} \; l_u ( \boldsymbol{x} )=0, \forall u; \; r_v ( \boldsymbol{x} )\leq0, \forall v. 
\end{aligned}
\label{eq:descent_direction_problem_sup}
\end{equation}

\noindent Note that since $\boldsymbol{x}_{k}$ is given in the $k$-th iteration and, therefore, is fixed, $I ( \boldsymbol{x}_{k})$ and the $\boldsymbol{a}_{i, k}$'s are also fixed. In the second stage, we prove that the $\hat{\boldsymbol{x}}$ described in the theorem is a solution to \pref{eq:descent_direction_problem_sup}, when $\nu > 1$. 

\noindent
\textbf{Stage 1}. We need to prove that the directional derivative of the objective function of \pref{eq:descent_direction_problem_sup} is negative, given the direction $\boldsymbol{d}_k = \hat{\boldsymbol{x}} - \boldsymbol{x}_k$. Since we only consider $\bar{g}$ to be affine in the individual entries of the kernel matrix, $g ( \boldsymbol{a}_{i, k}, \boldsymbol{x}_{k} )$ could be expressed as 

\begin{equation}
	g ( \boldsymbol{a}_{i, k}, \boldsymbol{x}_{k} ) = \sum_{t=1}^T  [  \sum_{m=1}^M \theta_{m,k}^t (\boldsymbol{b}_1^t(\boldsymbol{a}_{i, k})^{'}\hat{\boldsymbol{K}}_m^t  + b_2^t(\boldsymbol{a}_{i, k}) ) ] - \omega_{k}
\label{eq:proof_thm1_1}
\end{equation}

\noindent where $\boldsymbol{b}_1^t(\cdot)$ is a vector-valued function, $b_2^t(\cdot)$ is a scalar function and $\hat{\boldsymbol{K}}_m^t$ is a vector with elements $k_m(\boldsymbol{x}_i^t, \boldsymbol{x}_j^t ),$ $ i,j=1,\cdots,N_t$. If we define $I_+ ( \boldsymbol{x} ) $ $ \triangleq \{ i \in I ( \boldsymbol{x} ),$ $ g ( \boldsymbol{a}_i, \boldsymbol{x} ) > 0 \} $, $I_0 ( \boldsymbol{x} ) $ $  \triangleq \{ i \in I ( \boldsymbol{x} ), g ( \boldsymbol{a}_i, \boldsymbol{x} ) = 0 \} $, $I_- ( \boldsymbol{x} ) $ $  \triangleq  \{ i \in I  ( \boldsymbol{x} ), g ( \boldsymbol{a}_i, \boldsymbol{x} ) < 0 \} $, then the directional derivative can be calculated as 

\begin{equation}
	G(\boldsymbol{x}_{k}, \boldsymbol{d}_{k}, \nu)  = \boldsymbol{d}_{k}^{'} \phi (\boldsymbol{x}_{k}) + \nu \sum_{i \in I_{0}(\boldsymbol{x}_{k})} [\boldsymbol{d}_{k}^{'} 	\psi(\boldsymbol{a}_{i, k}, \boldsymbol{x}_{k})]_+  + \nu \sum_{i \in I_{+}(\boldsymbol{x}_{k})} \boldsymbol{d}_{k}^{'} \psi(\boldsymbol{a}_{i, k}, \boldsymbol{x}_{k})
\label{eq:proof_thm1_2}
\end{equation}

\noindent Let $\boldsymbol{d}_{k} \triangleq \hat{\boldsymbol{x}} - \boldsymbol{x}_{k}$, $H_{i,m,k}^t \triangleq \boldsymbol{b}_1^t(\boldsymbol{a}_{i, k})^{'}\hat{\boldsymbol{K}}_m^t + b_2^t(\boldsymbol{a}_{i, k})$ and $\boldsymbol{H}_{i,k}$ be the vector with elements $H_{i,m,k}^t$, $m=1,\cdots,M$, $t=1,\cdots,T$. Then, we obtain

\begin{equation}
	\begin{aligned}
	& G(\boldsymbol{x}_{k}, \boldsymbol{d}_{k}, \nu) = (\hat{\omega} - \omega_{k} ) \\
	& +\nu \sum_{i \in I_{0}(\boldsymbol{x}_{k})} [\hat{\boldsymbol{\theta}}^{'} \boldsymbol{H}_{i,k} - \hat{\omega} - \boldsymbol{\theta}_{k}^{'} \boldsymbol{H}_{i,k} + \omega_{k} ]_+ \\
	& + \nu \sum_{i \in I_{+}(\boldsymbol{x}_{k})} (\hat{\boldsymbol{\theta}}^{'} \boldsymbol{H}_{i,k} - \hat{\omega} - \boldsymbol{\theta}_{k}^{'} \boldsymbol{H}_{i,k} + \omega_{k} ) 
	\end{aligned}
\label{eq:proof_thm1_3}
\end{equation}

\noindent Note that $\boldsymbol{\theta}_{k}^{'} \boldsymbol{H}_{i,k} - \omega_{k} = g (\boldsymbol{a}_{i,k}, \boldsymbol{x}_{k}) = 0$, when $i \in I_{0}(\boldsymbol{x}_{k})$. Then, we get

\begin{equation}
	G(\boldsymbol{x}_{k}, \boldsymbol{d}_{k}, \nu) = f(\hat{\boldsymbol{x}}) + \nu \sum_{i \in I ( \boldsymbol{x}_{k} )} g ( \boldsymbol{a}_{i, k}, \hat{\boldsymbol{x}}  )_+  - f(\boldsymbol{x}_{k}) - \nu \sum_{i \in I ( \boldsymbol{x}_{k} )} g ( \boldsymbol{a}_{i, k}, \boldsymbol{x}_{k} )_+
\label{eq:fifteen}
\end{equation}

\noindent Since $\hat{\boldsymbol{x}}$ minimizes (\ref{eq:descent_direction_problem_sup}), we have that $G(\boldsymbol{x}_{k}, \boldsymbol{d}_{k}, \nu) \leq 0$.

\noindent
\textbf{Stage 2}. Denote the objective function in (\ref{eq:descent_direction_problem_sup}) as 
\begin{equation}
F( \omega, \boldsymbol{\theta} ) = F ( \boldsymbol{x} )=f ( \boldsymbol{x} )+\nu \sum_{i \in I( \boldsymbol{x}_{k})} g ( \boldsymbol{a}_{i, k}, \boldsymbol{x} )_+
\label{eq:objfun_descent_dir_sup}
\end{equation}

We first prove that in \pref{eq:descent_direction_problem_sup}, for $\nu > 1$ and fixed $\boldsymbol{\theta}$, the optimal solution to $\omega$ is $\hat{\omega}= \max_{i\in I ( \boldsymbol{x}_k )}$ $  \sum_{t=1}^T \bar{g}(\boldsymbol{\alpha}_{i,k}^t, \sum_{m=1}^M \theta_m^t \boldsymbol{K}_m^t)$, which yields $F ( \hat{\omega}, \boldsymbol{\theta} ) = \hat{\omega}$. To see that this is true, first consider another $\bar{\omega} > \hat{\omega}$, which implies that $F ( \bar{\omega}, \boldsymbol{\theta} ) = \bar{\omega} > \hat{\omega}$. On the other hand, if we consider $\bar{\omega} < \hat{\omega}$, then we have that 

\begin{equation}
	F ( \hat{\omega}, \boldsymbol{\theta} ) - F( \bar{\omega}, \boldsymbol{\theta} )  = \hat{\omega} - \bar{\omega} - \nu \sum_{i \in I ( \boldsymbol{x}_{k} )} (\sum_{t=1}^T \bar{g}(\boldsymbol{\alpha}_{i,k}^t, \sum_{m=1}^M \theta_m^t \boldsymbol{K}_m^t) -\bar{\omega} )_+
\label{eq:objfun_descent_sup_1}
\end{equation}

\noindent For simplicity, assume $\bar{\omega}$ is such that only one element in the summation over $i$ not zero. Then,

\begin{equation}
F ( \hat{\omega}, \boldsymbol{\theta} ) - F ( \bar{\omega}, \boldsymbol{\theta} ) = \hat{\omega} - \bar{\omega} - \nu (\hat{\omega} -\bar{\omega} )
\label{eq:objfun_descent_sup_2}
\end{equation}

\noindent Since $\nu > 1$, the above quantity is negative, which implies that $F ( \hat{\omega}, \boldsymbol{\theta} ) < F ( \bar{\omega}, \boldsymbol{\theta} )$. When $\bar{\omega}$ is such that more elements in the summation over $i$ are not zero, the rationale is similar. Therefore, we proved that for fixed $\boldsymbol{\theta}$, the previously defined $\hat{\omega}$ is the optimum.

Based on the just-stated fact, in order to minimize $F ( \omega, \boldsymbol{\theta} )$, we need to find a $\boldsymbol{\theta}$ such that $\max_{i\in I ( \boldsymbol{x} )} $ $ \sum_{t=1}^T \bar{g}(\boldsymbol{\alpha}_i^t, \sum_{m=1}^M \theta_m^t \boldsymbol{K}_m^t)$ is minimized:

\begin{equation}
\min_{\boldsymbol{\theta}}\max_{i\in I ( \boldsymbol{x}_k )}  \sum_{t=1}^T \bar{g}(\boldsymbol{\alpha}_{i,k}^t, \sum_{m=1}^M \theta_m^t \boldsymbol{K}_m^t)
\label{eq:objfun_descent_sup_3}
\end{equation}

\noindent This problem is equivalent to the following:

\begin{equation}
\begin{aligned} 
&\min_{\boldsymbol{\theta}} \lambda \\
& \text{s.t.} \sum_{t=1}^T \bar{g}(\boldsymbol{\alpha}_{i,k}^t, \sum_{m=1}^M \theta_m^t \boldsymbol{K}_m^t) \leq \lambda, \forall i \in I ( \boldsymbol{x}_k ) .
\end{aligned}
\label{eq:objfun_descent_sup_4}
\end{equation}

\noindent Therefore, we only need to solve the $\left | I_k \right |$ problems $\boldsymbol{\hat{\theta}}_i = \arg \min_{\boldsymbol{\theta}} \sum_{t=1}^T \bar{g}(\boldsymbol{\alpha}_{i,k}^t, \sum_{m=1}^M \theta_m^t \boldsymbol{K}_m^t), \forall i \in I ( \boldsymbol{x}_k )$, and find a value, say $\boldsymbol{\hat{\theta}}_{i_0}$, that minimizes the quantity $\max_{j \in I ( \boldsymbol{x}_k )} \{ \sum_{t=1}^T \bar{g} ( \boldsymbol{\alpha}_{j,k}^t, \sum_{m=1}^M \hat{\theta}_{m,i_0}^t\boldsymbol{K}_m^t )\}$.

\subsection{Proof to \propref{prop:closed-form_solution_group_sparsity}}
\label{sec:proof_to_closed-form_solution_group_sparsity}

The results for $p = 1, q = 1$ and $\boldsymbol{c}^t \succeq \boldsymbol{0}$ are obvious. In what follows, we assume that there is at least one negative element in each $\boldsymbol{c}^t$. We first prove the result for $p > 1, q > 1$. In this situation, the problem can be written as 

\begin{equation}
\begin{aligned}
\min_{\boldsymbol{\theta}\in \mathbb{R}^n, \boldsymbol{\sigma}\in \mathbb{R}^T} & \hspace{1mm} \boldsymbol{c}^\prime\boldsymbol{\theta}\\
\text{s.t.} \; & \boldsymbol{\theta} \succeq \boldsymbol{0}, \; \| \boldsymbol{\theta}^t \|_p = \sigma^t, \; \forall \; t;  \| \boldsymbol{\sigma} \|_q \leq a.
\end{aligned}
\label{eq:group_sparsity_sup_1}
\end{equation}

Note that the above problem is equivalent to the following:

\begin{equation}
\begin{aligned}
\min_{\boldsymbol{\theta}\in \mathbb{R}^n, \boldsymbol{\sigma}\in \mathbb{R}^T} & \hspace{1mm} \boldsymbol{c}^\prime\boldsymbol{\theta}\\
\text{s.t.} \; & \boldsymbol{\theta} \succeq \boldsymbol{0}, \;  \| \boldsymbol{\theta}^t \|_p \leq \sigma^t, \; \forall \; t;  \| \boldsymbol{\sigma} \|_q \leq a.
\end{aligned}
\label{eq:group_sparsity_sup_1_equiv}
\end{equation}

\noindent since the optimum $\boldsymbol{\theta}^t$ for the previously-stated problem must occur on the boundary. The Lagrangian of \pref{eq:group_sparsity_sup_1_equiv} takes the form of 

\begin{equation}
	L = \sum_{t=1}^T (\boldsymbol{c}^{t'} \boldsymbol{\theta}^t + \alpha^t ( \| \boldsymbol{\theta}^t \|_p - \sigma^t ) - \boldsymbol{\beta}^{t'} \boldsymbol{\theta}^t ) + \gamma ( \| \boldsymbol{\sigma} \|_q - 1 )
\label{eq:group_sparsity_sup_2}
\end{equation}

\noindent where $\alpha$ and $\boldsymbol{\beta}$ are the Lagrangian multipliers corresponding to the inequality constraints. Setting its partial derivatives with respect to $\boldsymbol{\theta}^t$ and $\boldsymbol{\sigma}$ to $\boldsymbol{0}$ and adding the complementary slackness conditions yields the following set of equalities: 

\begin{equation}
\left\{\begin{array}{l}
\boldsymbol{c}^t +\alpha^t  \| \boldsymbol{\theta}^t \|_p^{1-p} (\boldsymbol{\theta}^t)^{p-1} -\boldsymbol{\beta}^t = \boldsymbol{0}; \; \forall t \\ 
\gamma  \| \boldsymbol{\sigma} \|_q^{1-q} \boldsymbol{\sigma}^{q-1} -\boldsymbol{\alpha} = \boldsymbol{0};  \\
\alpha^t ( \| \boldsymbol{\theta}^t \|_p - \sigma^t ) = 0; \; \forall t\\
\beta_m^t \theta_m^t = 0; \; \forall m, \; t \\
\gamma (\| \boldsymbol{\sigma} \|_q - 1) = 0.
\end{array} \right.
\label{eq:A1} 
\end{equation}

\noindent Note that $\boldsymbol{\alpha}$ should not be $\boldsymbol{0}$, otherwise the minimum of $L$ would be either minus infinity or $0$, which are both trivial cases. Therefore, we have that

\begin{equation}
\left\{\begin{array}{l}
\| \boldsymbol{\theta}^t \|_p = \sigma^t; \; \forall t \\ 
\boldsymbol{\theta}^t = \sigma^t [\frac{1}{\alpha^t} ( \boldsymbol{\beta}^t - \boldsymbol{c}^t )]^{\frac{1}{p-1}}   \; \forall t
\end{array} \right.
\label{eq:A2} 
\end{equation}

\noindent Considering (\ref{eq:A1}) and the fact that $\alpha^t > 0$, as well as assuming that $\sigma^t > 0$, we have $\beta_m (\beta_m^t - c_m^t )^{\frac{1}{p-1}}=0, \; \forall m$. Since $\beta_i^t \geq  0$, we conclude that

%\begin{equation}
%\beta_m (\beta_m^t - c_m^t )^{\frac{1}{p-1}}=0, \; \forall m. 
%\label{eq:A3}
%\end{equation}

\begin{equation}
\left\{\begin{matrix}
\beta_m^t=0 & \text{if} \;c_m^t \leq  0\\ 
\beta_m^t=c_m^t & \text{otherwise}.
\end{matrix}\right.
\label{eq:A4}
\end{equation}

\noindent So, if we let $\tilde{\boldsymbol{c}}^t \triangleq [ \tilde{c}_1^t, \cdots, \tilde{c}_n^t ]^\prime$ and $\tilde{c}_m^t \triangleq \beta_m^t - c_m^t$, we obtain $\tilde{c}_m^t = \max \left \{ -c_m^t, 0 \right \}, \; \forall m$. We end up with $\boldsymbol{\theta}^t= \sigma^t (\frac{\tilde{\boldsymbol{c}}^t}{\alpha^t} )^{\frac{1}{p-1}} \propto ( \tilde{\boldsymbol{c}}^t )^\frac{1}{p-1}$. Since $\boldsymbol{\theta}^t$ needs to satisfy $\left \| \boldsymbol{\theta}^t \right \|_p = \sigma^t$, we normalize $ ( \tilde{\boldsymbol{c}}^t )^\frac{1}{p-1}$ and get 

%\begin{equation}
%\boldsymbol{\theta}^t= \sigma^t (\frac{\tilde{\boldsymbol{c}}^t}{\alpha^t} )^{\frac{1}{p-1}} \propto ( \tilde{\boldsymbol{c}}^t )^\frac{1}{p-1}
%\label{eq:A5}
%\end{equation}

\begin{equation}
\boldsymbol{\theta}^t = \sigma^t \frac{( \tilde{\boldsymbol{c}}^t)^{\frac{1}{p-1}}}{ \| \tilde{\boldsymbol{c}}^t \|_{\frac{p}{p-1}}^{\frac{1}{p-1}}}
\label{eq:A6}
\end{equation}

\noindent
In the previous derivation, we assumed that $\sigma^t \neq 0$. If $\sigma^t = 0$, we know immediately from (\ref{eq:A2}) that $\boldsymbol{\theta}^t = \boldsymbol{0}$. By the same rationale, we can show that $(\sigma^t)^{q-1} \propto  \alpha^t = \| \tilde{\boldsymbol{c}}^t \|_{\frac{p}{p-1}}$. Noting that the optimal $\boldsymbol{\sigma}$ must occur on the boundary, we arrive at the solution 

\begin{equation}
\sigma^t = \frac{a \| \boldsymbol{c}^t \|_{r+1}^s}{(\sum_{t=1}^T \| \boldsymbol{c}^t \|_{r+1}^{s+1})^{1/q}}.
\label{eq:A8} 
\end{equation}

When $q = 1$, (\ref{eq:A6}) is still valid. Therefore, we substitute (\ref{eq:A6}) into (\ref{eq:group_sparsity_sup_1_equiv}) to obtain:

\begin{equation}
\min_{\boldsymbol{\sigma}}  \sum_{t=1}^T \sigma^t  \| \tilde{\boldsymbol{c}}^t \|_{r+1} \; \text{s.t.} \; \boldsymbol{\sigma} > \boldsymbol{0}, \; \sum_{t=1}^t \sigma^t \leq a
\label{eq:A9} 
\end{equation}

\noindent Note that the constraint $\boldsymbol{\sigma} > \boldsymbol{0}$ is added, since $\sigma^t =  \| \boldsymbol{\theta}^t \|_p $. Obviously, the optimum $\boldsymbol{\sigma}$ is $\sigma^t = a[t = t_0]$, where $t_0 \triangleq \arg\min_{t}  \{  \| \hat{\boldsymbol{c}}^t \|_{r+1} \}$.

%\begin{equation}
%\sigma^t = a[t = \tilde{t}]
%\label{eq:A10} 
%\end{equation}

When $p = 1$, we first optimize with respect to $\boldsymbol{\theta}$. By fixing $\boldsymbol{\sigma}$, our problem can be split into $T$ minimization problems:

\begin{equation}
\min_{\boldsymbol{\theta}_t}  (\boldsymbol{c}^t)' \boldsymbol{\theta}^t \;\; \text{s.t.} \;\boldsymbol{\theta}^t \succeq \boldsymbol{0}, \sum_{i=1}^{N_t} \theta_i^t \leq \sigma^t
\label{eq:A11} 
\end{equation}

\noindent
The solution to this problem is $\boldsymbol{\theta}^{t*} = \sigma^t \boldsymbol{e}_{i_t}^t$. Substituting this result into \pref{eq:group_sparsity_sup_1_equiv}, optimizing with respect to $\boldsymbol{\sigma}$ and using an approach similar to the previous proof yields the desired result.

\end{document}